\documentclass[letterpaper, journal, onecolumn]{IEEEtran}
\usepackage{cite}
\usepackage{amsmath,amssymb,amsfonts}
\usepackage{algorithmic}
\usepackage{graphicx}
\usepackage{textcomp}
\usepackage{tabularx}  
\usepackage{ragged2e}  
\usepackage{makecell}  
\usepackage{array} 
\usepackage{mdframed}
\usepackage{comment}
\usepackage{url}
\usepackage[T1]{fontenc}

\newcolumntype{L}{>{\RaggedRight\arraybackslash}X} 

\def\BibTeX{{\rm B\kern-.05em{\sc i\kern-.025em b}\kern-.08em
    T\kern-.1667em\lower.7ex\hbox{E}\kern-.125emX}}

\title{\LARGE \bf
ChatGPT Empowered Long-Step Robot Control in Various Environments: A Case Application
}

\author{
Naoki Wake$^{1}$,
Atsushi Kanehira$^{1}$,
Kazuhiro Sasabuchi$^{1}$,
Jun Takamatsu$^{1}$,
and Katsushi Ikeuchi$^{1}$
\thanks{$^{1}$Applied Robotics Research, Microsoft, 
        Redmond, WA 98052, USA
        {\tt\small naoki.wake@microsoft.com}}%
}

\begin{document}
\maketitle
\thispagestyle{empty}
\pagestyle{empty}
\begin{abstract}
This paper introduces a novel method for translating natural-language instructions into executable robot actions using OpenAI's ChatGPT in a few-shot setting. We propose customizable input prompts for ChatGPT that can easily integrate with robot execution systems or visual recognition programs, adapt to various environments, and create multi-step task plans while mitigating the impact of token limit imposed on ChatGPT. In our approach, ChatGPT receives both instructions and textual environmental data, and outputs a task plan and an updated environment. These environmental data are reused in subsequent task planning, thus eliminating the extensive record-keeping of prior task plans within the prompts of ChatGPT. Experimental results demonstrated the effectiveness of these prompts across various domestic environments, such as manipulations in front of a shelf, a fridge, and a drawer. The conversational capability of ChatGPT allows users to adjust the output via natural-language feedback. Additionally, a quantitative evaluation using VirtualHome showed that our results are comparable to previous studies. Specifically, 36\% of task planning met both executability and correctness, and the rate approached 100\% after several rounds of feedback. Our experiments revealed that ChatGPT can reasonably plan tasks and estimate post-operation environments without actual experience in object manipulation. Despite the allure of ChatGPT-based task planning in robotics, a standardized methodology remains elusive, making our work a substantial contribution. These prompts can serve as customizable templates, offering practical resources for the robotics research community. Our prompts and source code are open source and publicly available at \url{https://github.com/microsoft/ChatGPT-Robot-Manipulation-Prompts}.
\end{abstract}

\maketitle
\begin{figure*}[ht]
  \centering
  \includegraphics[width=0.95\textwidth]{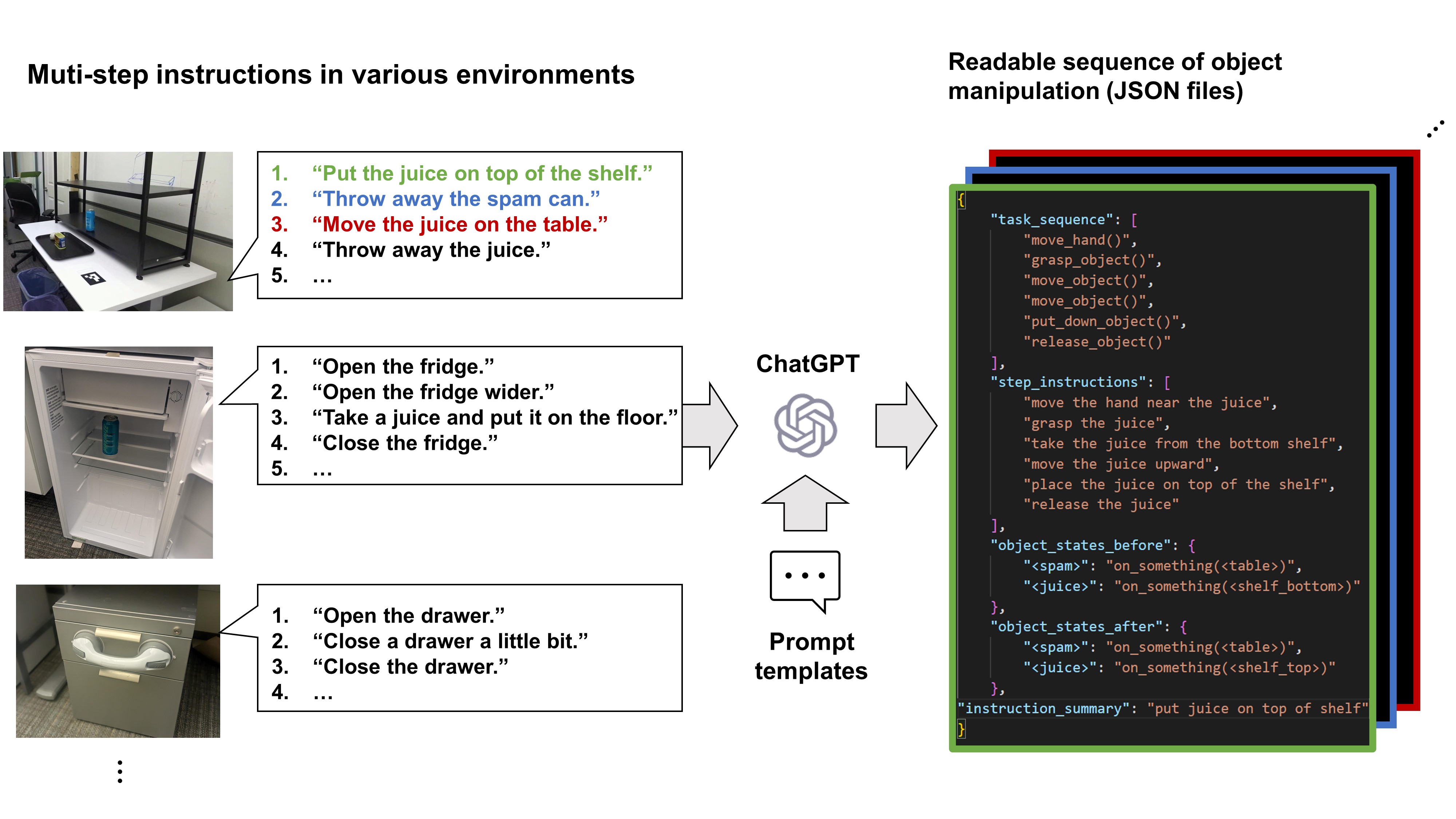}
  \caption{
  This paper presents practical prompts designed for various environments. The prompts enable ChatGPT to translate multi-step human instructions into sequences of executable robot actions.
  }
  \label{fig:gptexample}
\end{figure*}

\section{Introduction}
\label{sec:introduction}
Recent advances in natural language processing have yielded large language models (LLMs) with significantly improved abilities to understand and generate language. As a result of learning vast amounts of data, some LLMs can be fine-tuned given a small set of sample data as instructions (i.e., few-shot learning~\cite{brown2020language}). 
ChatGPT~\cite{OpenAI} is a representative example of such an LLM. One exciting application of ChatGPT is in the field of robotics, where it can be used for executable robot programs (i.e., task planning).

Task planning from natural-language instructions is a research topic in robotics, and there are many existing studies~\cite{pramanick2020decomplex, venkatesh2021translating, yanaokura2022multimodal}, some of which are built on top of LLMs~\cite{jiang2022vima, shridhar2023perceiver, brohan2023can, huang2022inner, ding2023task, singh2023progprompt,namasivayam2023learning, zhao2023differentiable,ding2022robot,zeng2022socratic,liang2023code,raman2022planning,xie2023translating} (\cite{Kovalev2022ApplicationOP} for review). However, most of them were developed within a limited scope of operations, such as pick-and-place~\cite{khan2023natural,kaynar2023remote,huang2022inner,zeng2022socratic}, are hardware-dependent, or lack the functionality of human-in-the-loop~\cite{skreta2023errors, huang2022language,ding2022robot, ding2023task,liang2023code}. Additionally, most of these studies rely on specific datasets~\cite{jiang2022vima,shridhar2023perceiver,lynch2020language,brohan2023can,pan2023dataefficient,lin2023text2motion,zhao2023erra,liu2023instructionfollowing,namasivayam2023learning,zhao2023differentiable,mees2023grounding}, which necessitate data recollection and model retraining when transferring or extending these to other robotic settings. 

In contrast to these pioneering studies, a significant advantage of utilizing most recent LLMs, such as ChatGPT, is their adaptability to various operational settings. This adaptability is facilitated by few-shot learning, which eliminates the need for extensive data collection or model retraining in customizing the scope of operations. Additionally, the recent LLMs' superior ability to process language allows for safe and robust task planning, as it can efficiently reflect user feedback in a human-in-the-loop manner.

In this study, we aim to demonstrate a specific but extensible use case of ChatGPT for task planning (Fig.~\ref{fig:gptexample}), employing ChatGPT as an example of the most recent LLMs. Although interest has been growing in the potential of ChatGPT in the field of robotics~\cite{vemprala2023chatgpt}, its practical application is still in its early stages, and no standardized methodology has yet been proposed. We design customizable prompts to meet the following requirements that are common to many practical robotic applications:
\begin{enumerate}
    \item Easy integration with robot execution systems or visual recognition programs.
    \item Applicability to various home environments.
    \item The ability to provide multi-step instructions while mitigating the impact of token limit imposed on ChatGPT.
\end{enumerate}
To meet these requirements, prompts are designed to have ChatGPT accomplish the following:
\begin{enumerate}
    \item Output a sequence of user-defined robot actions with explanations in an easy-to-parse JSON format.
    \item Explicitly handle the environmental information, enabling task planning considering the spatial relationships between objects.
    \item Estimate the post-operation environment as a hint for subsequent task planning, reducing the burden of holding lengthy conversation histories for multi-step instructions.
\end{enumerate}

Through experiments, we demonstrate that ChatGPT succeeds in estimating action sequences for multi-step instructions in various environments. Additionally, we show that the conversational capability of ChatGPT allows users to adjust the output through natural-language feedback, which is crucial for safe and robust task planning. Quantitative tests using the VirtualHome environment~\cite{puig2018virtualhome} show that the proposed prompts result in both executable and correct task planning after a few rounds of feedback, suggesting the effectiveness of our approach.

While previous research has validated the utility of ChatGPT within specific environments and scenarios~\cite{vemprala2023chatgpt}, we explored whether ChatGPT can operate effectively across diverse environments and scenarios. This attempt expands the practical applicability of ChatGPT, paving the way for broader and more flexible applications in various real-world settings. Our proposed prompts can serve as customizable templates and are open source and available at \url{https://github.com/microsoft/ChatGPT-Robot-Manipulation-Prompts}. Depending on the specifications of robot actions, environmental representations, and object names, users can easily modify them to meet their requirements. The contributions of this paper are threefold: we demonstrate the applicability of ChatGPT to multi-step task planning with a focus on robot action granularity, propose a customizable prompt adaptable to various environments, and make these prompts publicly accessible as a practical resource for the robotics research community.

\section{ChatGPT prompts}
The details of the designed prompts are explained in this section. The prompts consist of 1) an explanation of the role of ChatGPT, 2) a definition of robot actions, 3) an explanation of how to represent the environment, 4) an explanation of how to format the output, 5) examples of input and output, and 6) a specific instruction from the user.

In every instance of task planning with ChatGPT, the prompts one to five are loaded from pre-prepared text files, while the sixth prompt is dynamically generated based on the user's instructions and environmental information. Notably, through preliminary experimentation, we found that ChatGPT appears to operate more robustly when we input the six prompts as a conversation consisting of six turns (see Section~\ref{user_input} for details), rather than bundling them into a single prompt. All prompts and their output examples are available online \url{https://github.com/microsoft/ChatGPT-Robot-Manipulation-Prompts}, and anyone can try them out through OpenAI's API or a web browser.

The prompts shown in this section assumed that the robot has at least one arm, sufficient degrees of freedom, and reachability to execute the desired task in the working environment. Additionally, we assume that each instruction is given at the granularity of grasp-manipulation-release, which involves handling a single object from grasping to releasing. Challenges and discussions on extending our approach to more general-purpose robotic systems are discussed in Section \ref{discussion}.
\subsection{The role of ChatGPT}
In the first prompt, we provide ChatGPT with a context for this task by explaining the role that ChatGPT should play (Fig.~\ref{fig:p_role}). To accommodate multiple prompts, we include a sentence instructing ChatGPT to wait for the next prompt until all the prompts are input.

\begin{figure}[ht]
\begin{mdframed}[backgroundcolor=black]
\begin{flushleft}
\color[rgb]{0.7,0.7,0.7}\scriptsize
You are an excellent interpreter of human instructions for household tasks. Given an instruction and information about the working environment, you break it down into a sequence of robot actions. Please do not begin working until I say "Start working." Instead, simply output the message "Waiting for next input." Understood?
\end{flushleft}
\end{mdframed}
\caption{
The prompt for explaining the role of ChatGPT.
}
\label{fig:p_role}
\end{figure}

\subsection{The definition of robot actions}\label{action_definition}
In this prompt, we define a set of robot actions. Since an appropriate set of robot actions depends on the application and implementation of the robotic software, this prompt should be customized by experimenters. In Fig.~\ref{fig:p_functions}, we show an example of robot actions based on our in-house learning-from-observation application~\cite{wake2020learning, ikeuchi2021semantic}, in which robot actions are defined as functions that change the motion constraints on manipulated objects based on the Kuhn-Tucker theory~\cite{kuhn1956related}. This definition allows us to theoretically establish a necessary and sufficient set of robot actions for object manipulation. 
Experiments in Section~\ref{experiment} are conducted using these robot actions, except for an experiment in Section~\ref{virtualhome}, in which we defined a set of actions that were prepared for VirtualHome.

\begin{figure}[ht]
\begin{mdframed}[backgroundcolor=black]
\begin{flushleft}
\color[rgb]{0.7,0.7,0.7}\scriptsize
Necessary and sufficient robot actions are defined as follows:\\
\textquotedbl\textquotedbl\textquotedbl\\
"ROBOT ACTION LIST"\\
- move\_hand(): Move the robot hand from one position to another with/without grasping an object.\\
- grasp\_object(): Grab an object.\\
- release\_object(): Release an object in the robot hand.\\
...\\
- wipe\_on\_plane(): This action can only be performed if an object is grabbed. Move an object landing on a plane along two axes along that plane. For example, when wiping a window with a sponge, the sponge makes this motion.\\
\textquotedbl\textquotedbl\textquotedbl\\
\end{flushleft}
\end{mdframed}
\caption{
The prompt explaining a set of robot actions. See Fig.~\ref{fig:action_all} in Appendix~\ref{prompt} for the full action list. 
}
\label{fig:p_functions}
\end{figure}

\subsection{Representation of the environments}
This prompt defines the rule for representing working environments (Fig.~\ref{fig:p_env}). In this specific prompt, all physical entities are classified into non-manipulable obstacles, referred to as \textit{assets}, such as shelves and tables, and manipulable objects, referred to as \textit{objects}, such as cans and handles. These two classes are defined to differentiate between the entities that may be manipulated and those that cannot. As a hint for task planning, the spatial relationships between entities are described as \textit{states}, which are chosen from a ``STATE LIST.'' Through preliminary experimentation, items in the STATE LIST were identified as providing sufficient hints for ChatGPT to work effectively. Notably, the STATE LIST is customizable, and in Section~\ref{virtualhome}, we define different states to meet the specifications of VirtualHome. 

\begin{figure}[ht]
    \begin{mdframed}[backgroundcolor=black]
    \begin{flushleft}
    \color[rgb]{0.7,0.7,0.7}\scriptsize
    Information about environments and objects are given as Python dictionary. Example:\\
    \textquotedbl\textquotedbl\textquotedbl\\
    \{\hspace*{1em}\\
    \hspace*{2em}"environment":\{\hspace*{1em}\\
        \hspace*{4em}"assets": ["<table>", "<shelf\_bottom>", "<shelf\_top>", "<trash\_bin>", "<floor>"],\\
        \hspace*{4em}"asset\_states": \{"<shelf\_bottom>": "on\_something(<table>)", \\
        \hspace*{6em}"<trash\_bin>": "on\_something(<floor>)"\},\\
        \hspace*{4em}"objects": ["<spam>", "<juice>"],\\
        \hspace*{4em}"object\_states": \{"<spam>": "on\_something(<table>)", \\
        \hspace*{6em}"<juice>": "on\_something(<shelf\_bottom>)"\}\\
    \hspace*{2em}\}\\
    \}\\
    \textquotedbl\textquotedbl\textquotedbl\\
    Asset states and object states are represented using those state sets:\\
    \textquotedbl\textquotedbl\textquotedbl\\
    "STATE LIST"\\
    - on\_something(<something>): Object is located on <something>\\
    - inside\_something(<something>): Object is located inside <something>\\
    - inside\_hand(): Object is being grasped by a robot hand\\
    - closed(): Object can be opened\\
    - open(): Object can be closed or kept opened\\
    \textquotedbl\textquotedbl\textquotedbl\\
    <something> should be one of the assets or objects in the environment.
    \end{flushleft}
    \end{mdframed}
  \caption{
  The prompt for defining the rules for representing working environments.
  }
  \label{fig:p_env}
\end{figure}

\subsection{The format of the output produced by ChatGPT}
This prompt defines the format of the output produced by ChatGPT (Fig.~\ref{fig:p_format}). To facilitate easy integration with other pipelines, such as robot control systems and visual recognition programs, we encourage ChatGPT to output a Python dictionary that can be saved as a JSON file. Additionally, we encourage ChatGPT to include not only the sequence of robot actions, but also explanations of each action step and supplementary information on the updated environment after executing the actions. These additional pieces of information help the user debug whether ChatGPT correctly processes the input information. 

\begin{figure}[ht]
    \begin{mdframed}[backgroundcolor=black]
    \begin{flushleft}
    \color[rgb]{0.7,0.7,0.7}\scriptsize
    You divide the actions given in the text into detailed robot actions and put them together as a Python dictionary. The dictionary has five keys:\\
    \textquotedbl\textquotedbl\textquotedbl\\
    - dictionary["task\_cohesion"]: A dictionary containing information about the robot's actions that have been split up.\\
    - dictionary["environment\_before"]: The state of the environment before the manipulation.\\
    - dictionary["environment\_after"]: The state of the environment after the manipulation.\\
    - dictionary["instruction\_summary"]: contains a brief summary of the given sentence.\\
    - dictionary["question"]: If you cannot understand the given sentence, you can ask the user to rephrase the sentence. Leave this key empty if you can understand the given sentence.\\
    \textquotedbl\textquotedbl\textquotedbl\\
    Three keys exist in dictionary["task\_cohesion"].\\
    \textquotedbl\textquotedbl\textquotedbl\\
    - dictionary["task\_cohesion"]["task\_sequence"]: Contains a list of robot actions. Only the behaviors defined in the "ROBOT ACTION LIST" will be used.\\
    - dictionary["task\_cohesion"]["step\_instructions"]: contains a list of instructions corresponding to dictionary["task\_cohesion"]["task\_sequence"].\\
    - dictionary["task\_cohesion"]["object\_name"]: The name of the manipulated object. Only objects defined in the input dictionary will be used for the object name.\\
    \textquotedbl\textquotedbl\textquotedbl
    \end{flushleft}
    \end{mdframed}
  \caption{
  The prompt for defining the format of the output produced by ChatGPT.
  }
  \label{fig:p_format}
\end{figure}

\subsection{Examples of input and output}
This prompt provides examples of the expected inputs and outputs (Fig.~\ref{fig:p_example}). 
We found that providing more examples helps ChatGPT generate the desired sequence and thus minimizes the effort users need to expend to correct the output through conversations.

\begin{figure}[ht]
    \begin{mdframed}[backgroundcolor=black]
    \begin{flushleft}
    \color[rgb]{0.7,0.7,0.7}\scriptsize
    I will give you some examples of the input and the output you will generate.\\ 
    \textquotedbl\textquotedbl\textquotedbl\\
    Example 1:\\
    \textquotedbl\textquotedbl\textquotedbl\\
    - Input:\\
    \{\hspace*{1em}\\
    \hspace*{2em}...(environmental information)...\\
    \hspace*{2em}"instruction": "Put the juice on top of the shelf"\\
    \}\\
    - Output:\\
    \{\hspace*{1em}\\
    \hspace*{2em}"task\_cohesion": \{\hspace*{1em}\\
    \hspace*{4em}"task\_sequence": [\\
    \hspace*{6em}"move\_hand()",\\
    \hspace*{6em}"grasp\_object()",\\
    \hspace*{6em}... ,\\
    \hspace*{6em}"attach\_to\_plane()",\\
    \hspace*{6em}"release\_object()"\\
    \hspace*{4em}],\\
    \hspace*{4em}"step\_instructions": [\\
    \hspace*{6em}"move the hand near the juice",\\
    \hspace*{6em}"grasp the juice",\\
    \hspace*{6em}... ,\\
    \hspace*{6em}"place the juice",\\
    \hspace*{6em}"release the juice"\\
    \hspace*{4em}],\\
    \hspace*{4em}"object\_name": "<juice>"\\
    \hspace*{2em}\},\\
    \hspace*{2em}"environment\_before": ... ,\\
    \hspace*{2em}"environment\_after": ... ,\\
    \hspace*{2em}"instruction\_summary": "put the juice on top of the shelf",\\
    \}\\
    \textquotedbl\textquotedbl\textquotedbl\\
    Example 2:\\
    \textquotedbl\textquotedbl\textquotedbl\\
    ...\\
    \end{flushleft}
    \end{mdframed}
  \caption{
  The prompt providing examples of desired inputs and outputs. The full information is available at the URL provided in the text.
  }
  \label{fig:p_example}
\end{figure}

\subsection{Specific instruction from the user}\label{user_input}
\begin{figure*}[ht]
  \centering
  \includegraphics[width=0.6\textwidth]{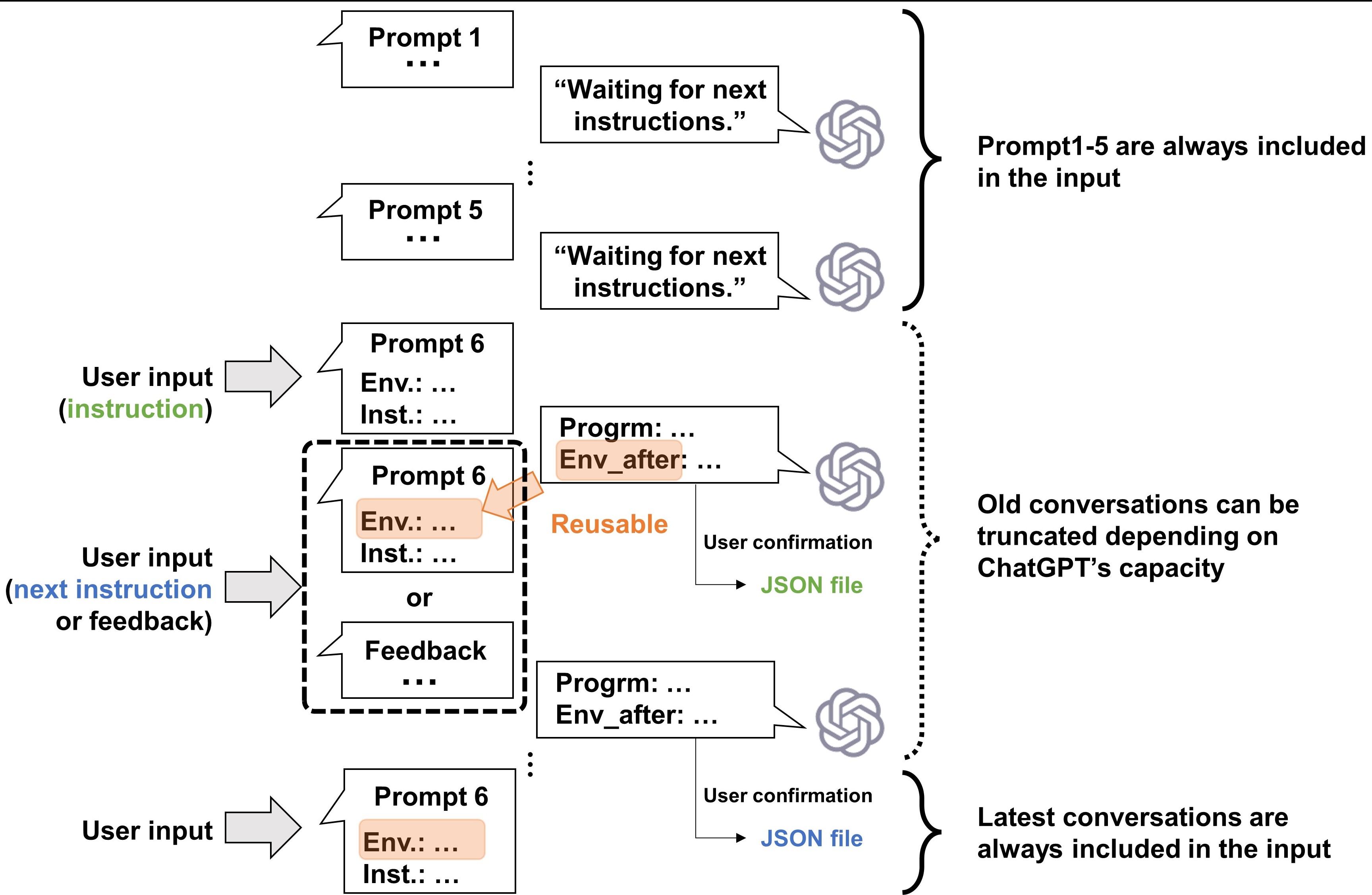}
  \caption{
  The entire structure of the conversation with ChatGPT in task planning.
  }
  \label{fig:p_structure}
\end{figure*}

While the previous five prompts are fixed, the sixth prompt is dynamically generated in every instance of task planning by editing a template prompt (Fig.~\ref{fig:p_query}). This prompt is generated by replacing [INSTRUCTION] with the given instruction and [ENVIRONMENT] with the corresponding environmental information. Notably, the user is required to provide the environmental information in the initial instance of task planning using a separate process (e.g., manual preparation). However, this effort is unnecessary for subsequent instances because we can reuse an updated environment incorporated in the last output of ChatGPT (orange parts in Fig.~\ref{fig:p_structure}). This approach facilitates task planning based on the most recent environment, eliminating the need for extensive historical records that exceed ChatGPT's token limit. In our experiments, we practically included as much of the historical record as the token limit of ChatGPT would allow, ranging from the most recent to the oldest conversation history (Fig.~\ref{fig:p_structure}).

As a specific usage of this task planner within a robot system, we assume that the output of ChatGPT is checked by the user in every instance of task planning. If the user confirms that there is no further need for adjustment, the output is then saved as a JSON file. In Appendix~\ref{appendix_robot}, we have provided more details regarding how the proposed task planner is integrated and operated within a robot system.

\begin{figure}[ht]
    \begin{mdframed}[backgroundcolor=black]
    \begin{flushleft}
    \color[rgb]{0.7,0.7,0.7}\scriptsize
    \textquotedbl\textquotedbl\textquotedbl\\
    \{"environment":[ENVIRONMENT]\}\\
    \textquotedbl\textquotedbl\textquotedbl\\
    The instruction is as follows:\\
    \textquotedbl\textquotedbl\textquotedbl\\
    \{"instruction": [INSTRUCTION]\}\\
    \textquotedbl\textquotedbl\textquotedbl\\
    The dictionary that you return should be formatted as python dictionary. Follow these rules:\\
    1. The first element should be move\_hand() to move the robot hand closer to the object. Always end with releasing the object.\\  
    2. Make sure that each element of the ["step\_instructions"] explains corresponding element of the ["task\_sequence"]. Refer to the "ROBOT ACTION LIST" to understand the elements of ["task\_sequence"].\\
    ...\\
    9. Make sure that you output a consistent manipulation as a single arm robot. For example, grasping an object should not occur in successive steps.\\
    Adhere to the output format I defined above. Follow the nine rules. Think step by step.
    \end{flushleft}
    \end{mdframed}
  \caption{
  The user input template and examples of the actual input used. The user is assumed to provide environmental information. Multi-step task planning can be realized by reusing the environmental information that ChatGPT outputs in the following task planning.
  }
  \label{fig:p_query}
\end{figure}

\section{Experiments}\label{experiment}
We tested the prompts to verify whether ChatGPT behaves in accordance with the specified requirements. We used a fixed GPT model provided by Azure OpenAI (gpt-3.5-turbo) in our experiments. Some experimental results are not fully presented in order to save space, but all results, including parameters for ChatGPT inference, instructions, and environment definitions, can be found here: \url{https://github.com/microsoft/ChatGPT-Robot-Manipulation-Prompts}.

\subsection{Multi-step manipulation of the environment}
We tested the applicability of the proposed prompts to multi-step instructions in various environments. As examples of household tasks, we conducted role-plays instructing the rearrangement and disposal of objects placed on tables and shelves, retrieving objects from refrigerators and drawers, and cleaning tables and windows with a sponge. The instructions and feedback texts were prepared in a style that resembles the way humans communicate with each other. The environmental information in the initial instance of task planning was prepared manually for each scenario. The output of ChatGPT was manually checked by the authors at every instruction step. 
Specifically, we conducted a visual inspection to qualitatively confirm whether the generated action sequences were executable and whether they accompanied reasonable environment estimations. 
In summary, the results shown below suggest that ChatGPT can translate multi-step human instructions into adequate sequences of executable robot actions.

\subsubsection{Relocation of objects on a table}
\begin{figure*}[ht]
  \centering
  \includegraphics[width=1.0\textwidth]{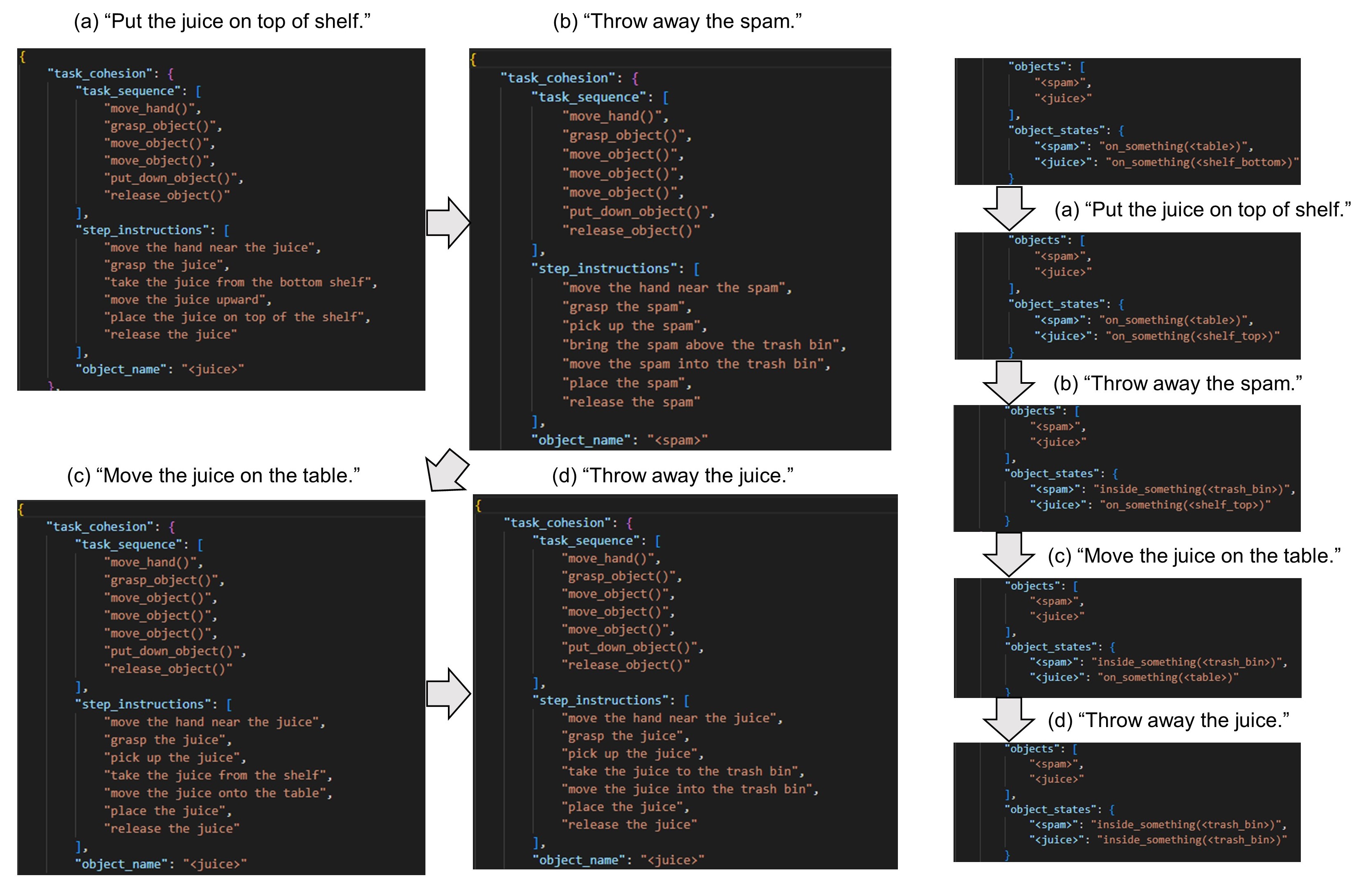}
  \caption{
  An example of the output produced by ChatGPT for the task of relocating objects. (Left panel) Robot actions broken down for each natural language. (Right panel) The state of the environment that is output by ChatGPT. A part of the JSON output is shown for each file. All the results, including the representation of the environment can be found here: https://github.com/microsoft/ChatGPT-Robot-Manipulation-Prompts.
  }
  \label{fig:shelf}
\end{figure*}

The task involves manipulating a can of juice situated on the bottom shelf of a two-shelf structure and a can of spam positioned on a table (refer to the top panel in Fig.~\ref{fig:gptexample} for the scene). First, the juice is relocated from the bottom to the top shelf. Subsequently, the spam is discarded into a trash bin. Thereafter, the juice is moved from the top shelf to the table. Finally, the juice, too, is discarded into the trash bin. The output of ChatGPT, which demonstrates successful task planning, is shown in Fig.~\ref{fig:shelf}.

\subsubsection{Open a fridge/drawer door}
\begin{figure*}[ht]
  \centering
  \includegraphics[width=1.0\textwidth]{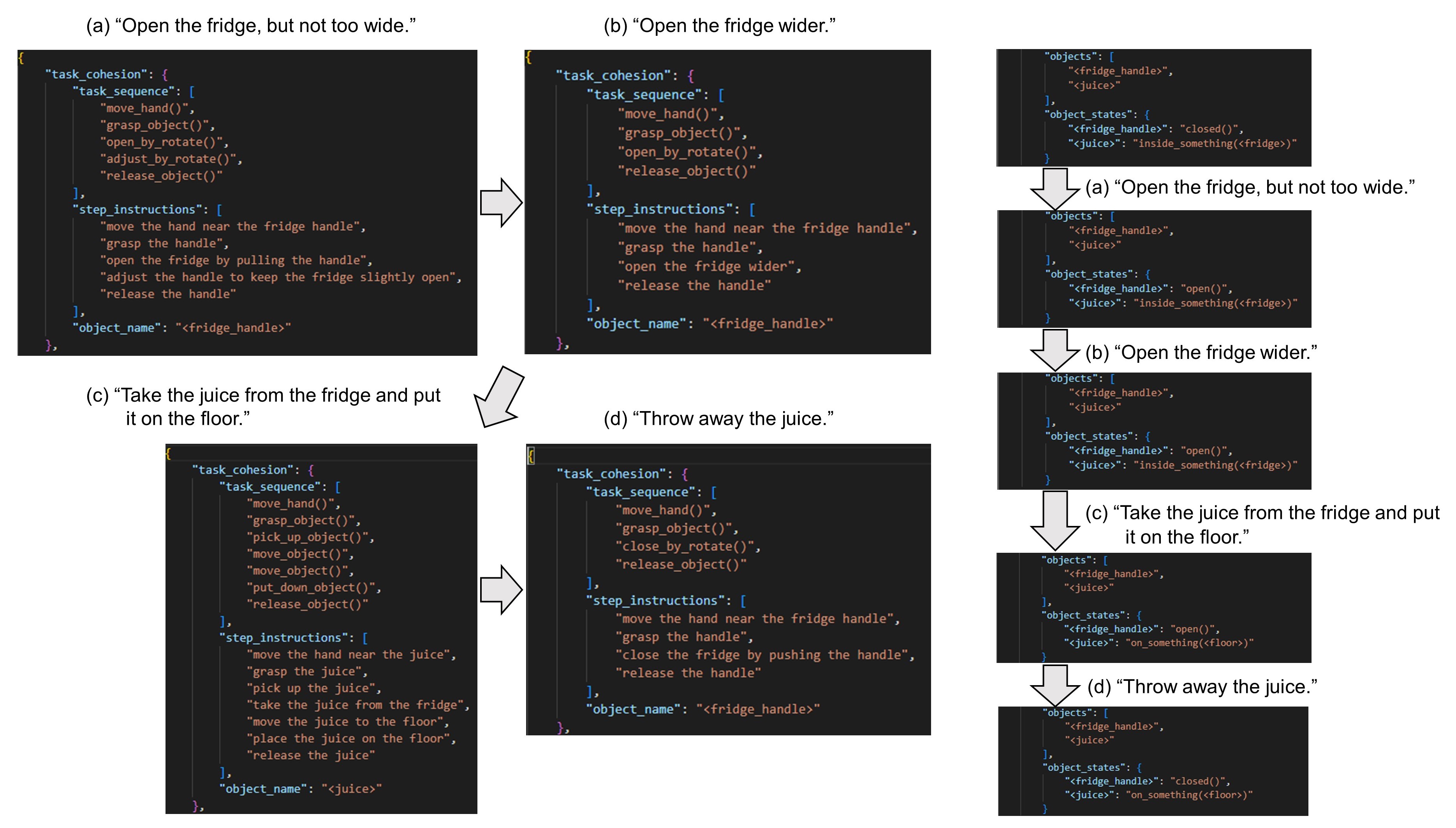}
  \caption{
  An example of the output produced by ChatGPT for the task of opening a refrigerator and retrieving juice. (Left panel) Robot actions broken down for each natural language. (Right panel) The state of the environment that is output by ChatGPT. A part of the JSON output is shown for each file. All the results, including the representation of the environment can be found here: https://github.com/microsoft/ChatGPT-Robot-Manipulation-Prompts.
  }
  \label{fig:fridge}
\end{figure*}

Next, we tested the scenario of opening a refrigerator door, opening the door slightly wider, removing a juice from the refrigerator and placing it on the floor, and finally closing the refrigerator (see the middle panel in Fig.~\ref{fig:gptexample} for the scene). The output of ChatGPT is shown in Fig.~\ref{fig:fridge}, indicating a successful task planning. Similar results were obtained for the scenario of sliding a drawer open (Figure not shown).

\subsubsection{Wipe a window with a sponge, and throw it away}
\begin{figure*}[ht]
  \centering
  \includegraphics[width=1.0\textwidth]{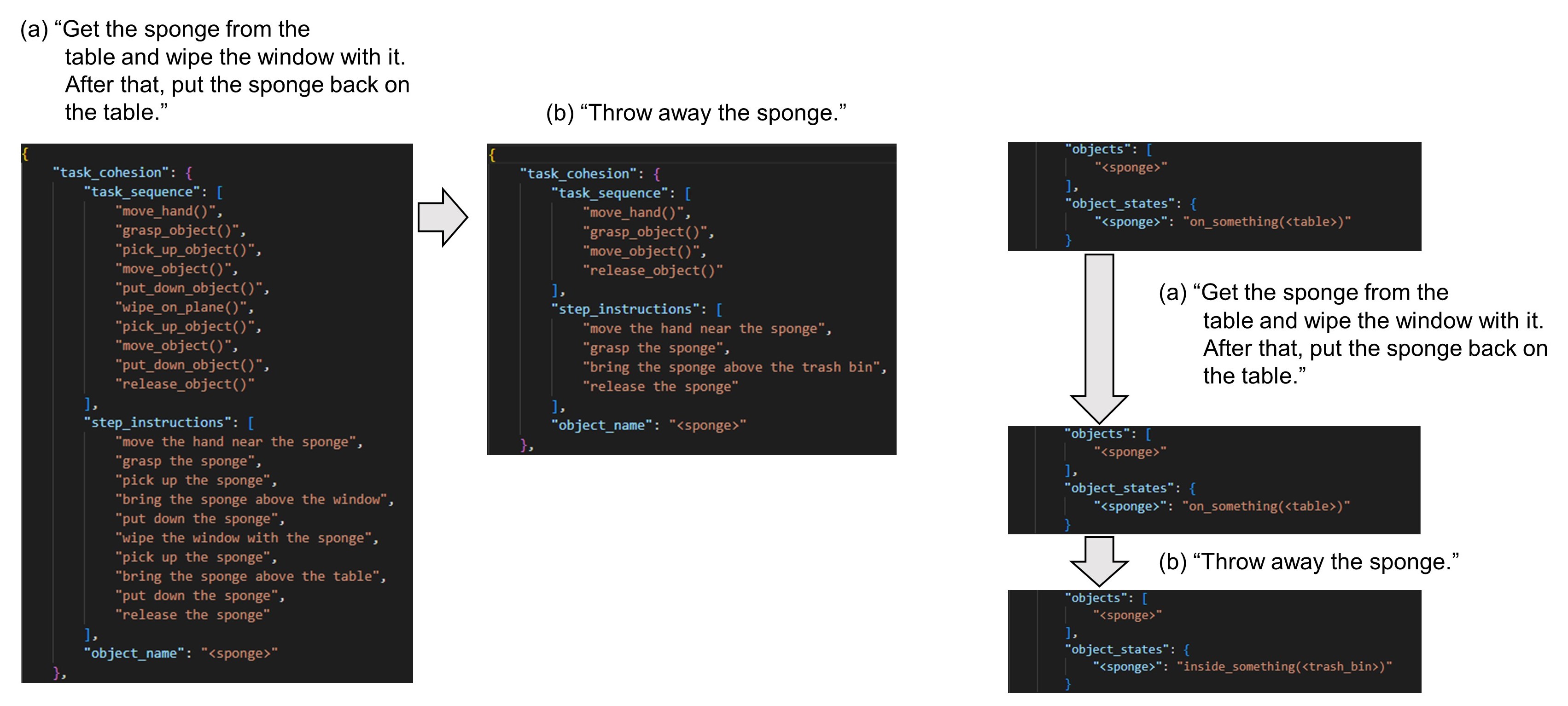}
  \caption{
  An example of the output produced by ChatGPT for the task of wiping a window with a sponge (Left panel) Robot actions broken down for each natural language. (Right panel) The state of the environment that is output by ChatGPT. A part of the JSON output is shown for each file. All the results, including the representation of the environment can be found here: https://github.com/microsoft/ChatGPT-Robot-Manipulation-Prompts.
  }
  \label{fig:window}
\end{figure*}

Next, we tested the scenario of taking a sponge from the desk, wiping the window with the sponge, and returning it to the table. Following the operation, a user throws the sponge into a trash bin. The output of ChatGPT is shown in Fig.~\ref{fig:window}, indicating a successful task planning. Similar results were obtained for the scenario of wiping the table with a sponge (data not shown).

\subsection{Adjustment of the output produced by ChatGPT through user feedback}\label{correction}
Since ChatGPT does not always generate complete action sequences, it is important for users to review and correct errors to ensure safe and robust operation. With this in mind, we tested the ability of ChatGPT to adjust the output through natural-language feedback.

\begin{figure}[ht]
  \centering
  \includegraphics[width=0.48\textwidth]{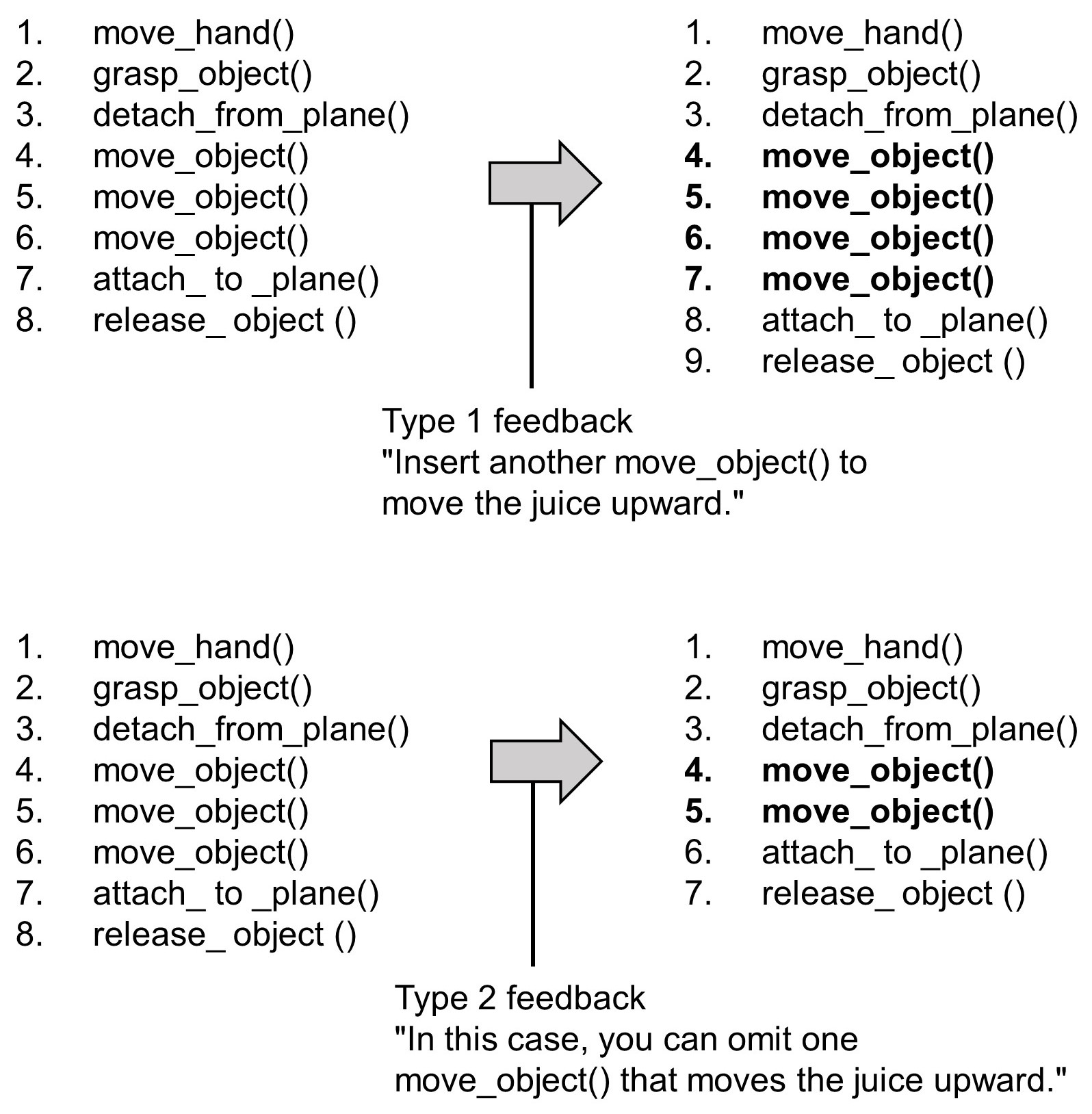}
  \caption{
    An example of adjusting an output sequence through natural-language feedback. The initial instruction was to move a juice from the bottom shelf to the top shelf. (Top panel) After the feedback of ``Insert another move\_object() to move the juice upward.,'' an action of move\_object() was added to the sequence. (Bottom panel) After the feedback of ``In this case, you can omit one move\_object() that moves the juice upward,'' an action of move\_object() was deleted from the sequence.
}
  \label{fig:addition_deletion}
\end{figure}

Fig.~\ref{fig:addition_deletion} shows the result when a user asked ChatGPT to add/remove a task in the output sequence. ChatGPT changed the output following the semantic content of the feedback, suggesting the functionality for making the necessary adjustments.

\subsection{Quantitative evaluation of task decomposition performance of ChatGPT}\label{virtualhome}
The previous sections qualitatively demonstrated that the proposed prompts achieve successful task planning using an action set from our in-house learning-from-observation system. In this section, we quantitatively evaluate the performance of task planning using a general-use simulation environment called VirtualHome~\cite{puig2018virtualhome}. Specifically, we made ChatGPT generate task plans from a single instruction for several household operation scenarios, and tested whether the resulting action sequences were valid in terms of executability in simulation and correctness upon visual inspection. All the source codes and prompts used for the experiment can be found here: \url{https://github.com/microsoft/ChatGPT-Robot-Manipulation-Prompts}.

\subsubsection{Experimental setup}
VirtualHome is software that simulates interactions between an agent and various home environments. The agent can navigate and interact within these environments by executing sequences of commands using a Python-based API. This API provides a set of pre-defined atomic actions (see Table~\ref{tab:human_action_list} in Appendix~\ref{appendix} for the action list), which represent the smallest units of action in VirtualHome. The environment consists of typical household objects (e.g., a plate, a microwave, and a table), each associated with unique IDs. The relationships between objects are represented as a graph that can be accessed through the API.

In the experiments, we selected a kitchen as a representative home environment and defined test scenarios for household chores that could be implemented in VirtualHome. To mitigate bias in the process of scenario preparation, we utilized ChatGPT to generate scenario candidates\footnote{ChatGPT was provided with the list of the kitchen objects and the pre-defined atomic actions to generate the candidates.}. We then manually selected the test scenarios that satisfied the following criteria:
\begin{itemize}
    \item The scenario can be realized by executing multiple actions in sequence.
    \item The scenario involves at least one instance of object manipulation, i.e., grasping or releasing an object.
    \item The tasks in the scenario are relevant to everyday activities.
\end{itemize}

Fourteen scenarios were prepared as test scenarios (Table~\ref{table:scenarios}). For these scenarios, we manually identified action sequences to achieve the scenarios along with the list of objects involved with their IDs. The action sequences identified are provided in Table
\ref{table:scenarioswithactions} in Appendix~\ref{appendix}.
\begin{table}[ht]
\caption{The list of scenarios used in the experiment}
\centering
\renewcommand\arraystretch{1.5} 
\begin{tabularx}{0.49\textwidth}{|c|X|}
\hline
\textbf{Scenario} & \textbf{Textual instruction} \\
\hline
Scenario1 & Take the bread from the toaster on the kitchen counter and place it on the plate on the table. \\
\hline
Scenario2 & Take the frying pan from the counter and place it in the sink. \\
\hline
Scenario3 & Take the pie from the table and put it in the microwave. \\
\hline
Scenario4 & Take the condiment shaker from the bookshelf and place it on the table. \\
\hline
Scenario5 & Take the book from the table and put it on the bookshelf. \\
\hline
Scenario6 & Take the water glass from the table and drink from it. \\
\hline
Scenario7 & Take the salmon on top of the microwave and put it in the fridge. \\
\hline
Scenario8 & Turn on the TV. \\
\hline
Scenario9 & Put a plate that is on the table into the sink. \\
\hline
Scenario10 & Take the pie on the table and warm it using the stove. \\
\hline
Scenario11 & Put the sponge in the sink and wet it with water. \\
\hline
Scenario12 & Take the knife from the table and move it to another place on the table. \\
\hline
Scenario13 & Take the plate from the table and move it to another place on the table. \\
\hline
Scenario14 & Take the condiment bottle from the bookshelf and put it on the table. \\
\hline
\end{tabularx}
\label{table:scenarios}
\end{table}

ChatGPT generated an action sequence intended to complete the scenario, given the proposed prompts, environmental information, and an instruction provided in the right column of Table~\ref{table:scenarios}. The environmental information, which corresponds to each scenario, was derived from the graph. Because of redundancy in representing all kitchen objects, only those objects involved in each scenario were considered. The action sequence generated by ChatGPT was then converted into a format that VirtualHome could interpret and executed in a step-by-step manner through the API. 
An action sequence was considered successful when the following two conditions were met:
\begin{itemize}
    \item Executability: The simulator was able to execute all steps without encountering any errors.
    \item Correctness: Upon visual inspection, it was determined that the proposed action steps could successfully complete the scenario.
\end{itemize}
We incorporated visual inspections in our criteria because a successful execution in the simulator does not necessarily guarantee that the final goal is achieved~\cite{huang2022language}. 

\subsubsection{Results}
We first tested whether the generated action sequences were successful without feedback. Since we conducted multiple trials, we set the temperature parameter to its maximum to ensure trial-to-trial variations in the output of ChatGPT. Table~\ref{table:result_vh} shows the results, with a success rate of approximately 36\% (5 out of 14 scenarios), with only a minimal variation observed between trials.

\begin{table*}[ht]
\caption{Executability of the output action sequence across trials. ``1'' indicates success, and ``0'' indicates failure.}
\centering
\begin{tabular*}{0.7\textwidth}{@{\extracolsep{\fill}}|c|c|c|c|c|c|c|c|c|c|c|c|c|c|c|}
\hline
Scenario & 1 & 2 & 3 & 4 & 5 & 6 & 7 & 8 & 9 & 10 & 11 & 12 & 13 & 14 \\
\hline
Trial 1 & 0 & 0 & 0 & 0 & 1 & 1 & 0 & 1 & 0 & 0 & 0 & 1 & 1 & 0 \\
Trial 2 & 0 & 0 & 0 & 0 & 1 & 1 & 0 & 1 & 0 & 0 & 0 & 1 & 1 & 0 \\
Trial 3 & 0 & 0 & 0 & 0 & 1 & 1 & 0 & 1 & 0 & 0 & 0 & 1 & 1 & 0 \\
Trial 4 & 0 & 0 & 0 & 0 & 1 & 1 & 0 & 1 & 0 & 0 & 0 & 1 & 1 & 0 \\
Trial 5 & 0 & 0 & 0 & 0 & 1 & 1 & 0 & 1 & 0 & 0 & 0 & 1 & 1 & 0 \\
\hline
\end{tabular*}
\label{table:result_vh}
\end{table*}

Upon investigating the unsuccessful cases, we identified two failure patterns in ChatGPT:
\begin{itemize}
    \item Incorrect verb selection: In VirtualHome, the simulator raises errors when it fails to select an action applicable to an object. For example, when the task involves `placing an object,' the action `PutIn' should be selected when placing the object inside a container, while the action `Put' should be selected when placing it on a flat surface. Despite these verb selection rules being part of the prompts, ChatGPT sometimes confused the actions.
    \item Omission of necessary steps: Some outputs skipped essential steps necessary for successfully completing a scenario, such as opening a container before placing an object inside it.
\end{itemize}

Following this analysis, we investigated whether adjustments could be made with a reasonable amount of effort, given appropriate feedback from a user who is familiar with task planning. To this end, we prepared an automatic feedback system as an objective method that detects these types of errors. This system was designed to simulate a user who is knowledgeable in task planning, instead of relying on manual feedback. The output of ChatGPT was checked by the system in every instance of task planning, and if an error was detected, an error message was automatically generated and fed back to ChatGPT. Using this feedback system, we examined the number of rounds of feedback needed to reach a successful sequence, or whether it was possible at all, across 14 scenarios. We set the temperature parameter to its minimum to ensure the reproducibility of the results.

\begin{figure*}[ht]
  \centering
  \includegraphics[width=0.8\textwidth]{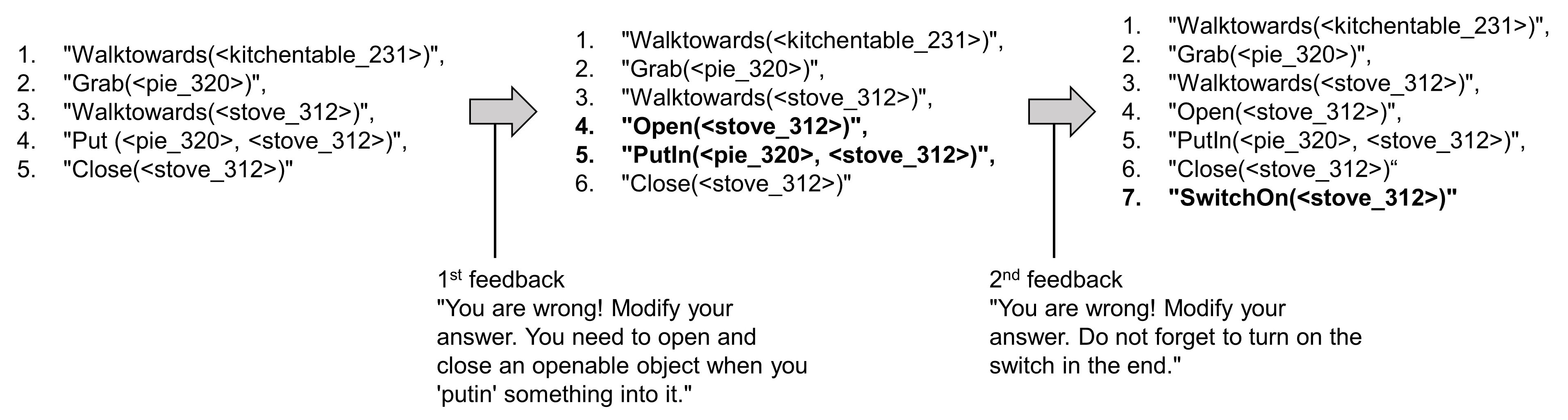}
  \caption{
  Example of adjusting an output sequence through auto-generated feedback. The output for the scenario 10 is shown (i.e., Take the pie on the table and warm it using the stove.) All the results, including the representation of the environment, can be found here: https://github.com/microsoft/ChatGPT-Robot-Manipulation-Prompts.
  }
  \label{fig:vh_correcion}
\end{figure*}

Table~\ref{table:result_vh_feedback} shows the results. ChatGPT was able to produce successful action sequences in all scenarios after receiving several rounds of feedback. Fig.~\ref{fig:vh_correcion} shows an example where auto-generated feedback texts guide ChatGPT towards a successful action sequence, suggesting that ChatGPT is capable of reflecting the semantic content of the feedback in its output and making the necessary adjustments.

\begin{table*}[ht]
\caption{The number of rounds of feedback needed to reach a successful sequence}
\centering
\begin{tabular*}{0.7\textwidth}{@{\extracolsep{\fill}}|c|c|c|c|c|c|c|c|c|c|c|c|c|c|c|}
\hline
Scenario & 1 & 2 & 3 & 4 & 5 & 6 & 7 & 8 & 9 & 10 & 11 & 12 & 13 & 14 \\
\hline
Number of feedback & 1 & 1 & 3 & 1 & 0 & 0 & 1 & 0 & 1 & 2 & 1 & 0 & 0 & 1 \\
\hline
\end{tabular*}
\label{table:result_vh_feedback}
\end{table*}

Our proposed prompts aim to estimate the post-operation environment as a hint for subsequent task planning, which enables multi-step task planning beyond the token limit imposed on ChatGPT. Therefore, we visually inspected the output of ChatGPT in Table~\ref{table:result_vh} and examined whether the post-operation environment was accurately estimated. As a result, we found that for all 14 scenarios and five trials, the estimation by ChatGPT was accurate, regardless of the success or failure of the action sequence. This result suggests that our proposed prompts can be adopted for multi-step task planning beyond a single scenario. The results can be found here: \url{https://github.com/microsoft/ChatGPT-Robot-Manipulation-Prompts}.

\section{Discussion: towards more general robotic applications}\label{discussion}
In this study, we focused on task planning of robot actions from multi-step instructions. We designed prompts for ChatGPT to meet three requirements: 1) easy integration with robot execution systems or visual recognition programs, 2) applicability to various environments, and 3) the ability to provide multi-step instructions while mitigating the impact of token limit imposed on ChatGPT. Through experiments, we confirmed that the proposed prompts work for multi-step instructions in various environments, and that ChatGPT enables the user to adjust the output interactively. Based on these results, we believe that the proposed prompts are practical resources that can be widely used in the robotics research community.

It is noteworthy that ChatGPT is capable of performing task planning without any actual experience in object manipulation, relying solely on few-shot data. This ability may be attributed to the fact that the model acquires knowledge of object manipulation and the temporal relationships between cohesively occurring actions during its training on a vast amount of data. In fact, the ability of ChatGPT to generate recipes from a menu suggests that it implicitly learns procedural steps~\cite{vemprala2023chatgpt}. Nevertheless, we cannot access the inner computations of ChatGPT, thus the computation process for task planning, including the estimation of the post-operation environment, remains unclear.

The quantitative analysis using VirtualHome showed that ChatGPT produced action sequences with a success rate of 36\% without feedback, which approached 100\% after several rounds of feedback. Although we used a different task set, our results align with those of previous studies that used LLMs for task planning in VirtualHome. Huang et al.~\cite{huang2022language} reported that 35.23\% of an LLM's outputs were both executable and correct from a human perspective. Raman et al.~\cite{raman2022planning} showed an improvement in both task execution and correctness through re-prompting based on precondition error information. Thus, we emphasize that our findings attest to the effectiveness of the proposed prompts, which aligns with the existing research.

The prompts were designed under the assumption that the robot has at least one arm, sufficient degrees of freedom, and reachability to execute the desired task in a given environment. Additionally, we assume that each instruction is given at the granularity of grasp-manipulation-release. However, these assumptions may be restrictive for some scenarios in general robotic manipulations. In the following sections, we discuss several strategies to effectively integrate our task planner with practical robotic applications.

\subsection{Handling of conditional branching}
\begin{figure}[ht]
  \centering
  \includegraphics[width=0.48\textwidth]{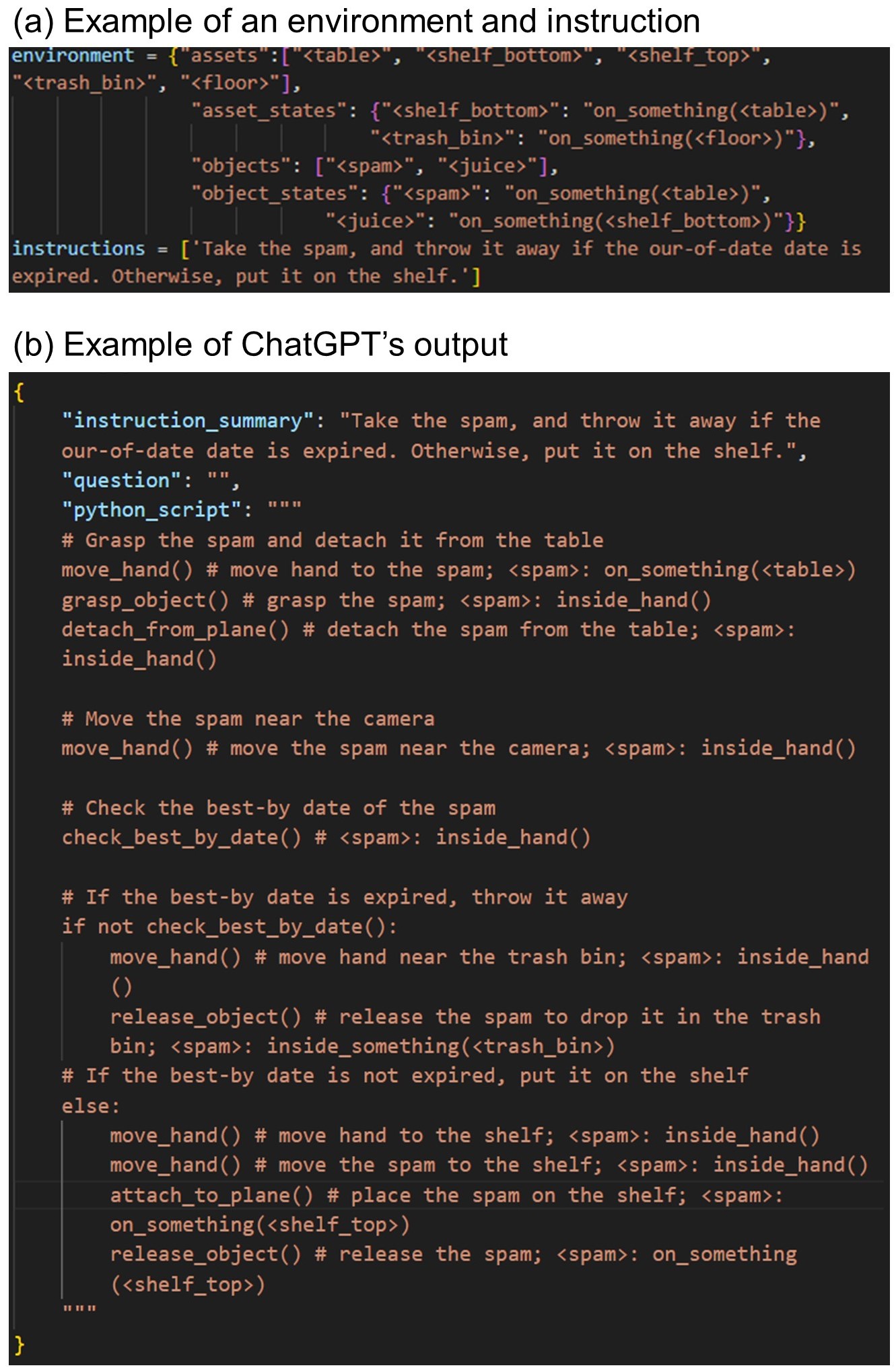}
  \caption{
  An example demonstrating the feasibility of using ChatGPT to generate control programs that include conditional branching. A part of the prompts is shown. Note that we encouraged ChatGPT to add comments at every line to track the state of objects, as the final state may vary according to the conditional branching. We also added a non-manipulative function (i.e., check\_best\_by\_date()) in the robot action set. All the results, including the representation of the environment, can be found here: https://github.com/microsoft/ChatGPT-Robot-Manipulation-Prompts.
  }
  \label{fig:logic_prompt}
\end{figure}

\begin{figure}[ht]
  \centering
  \includegraphics[width=0.48\textwidth]{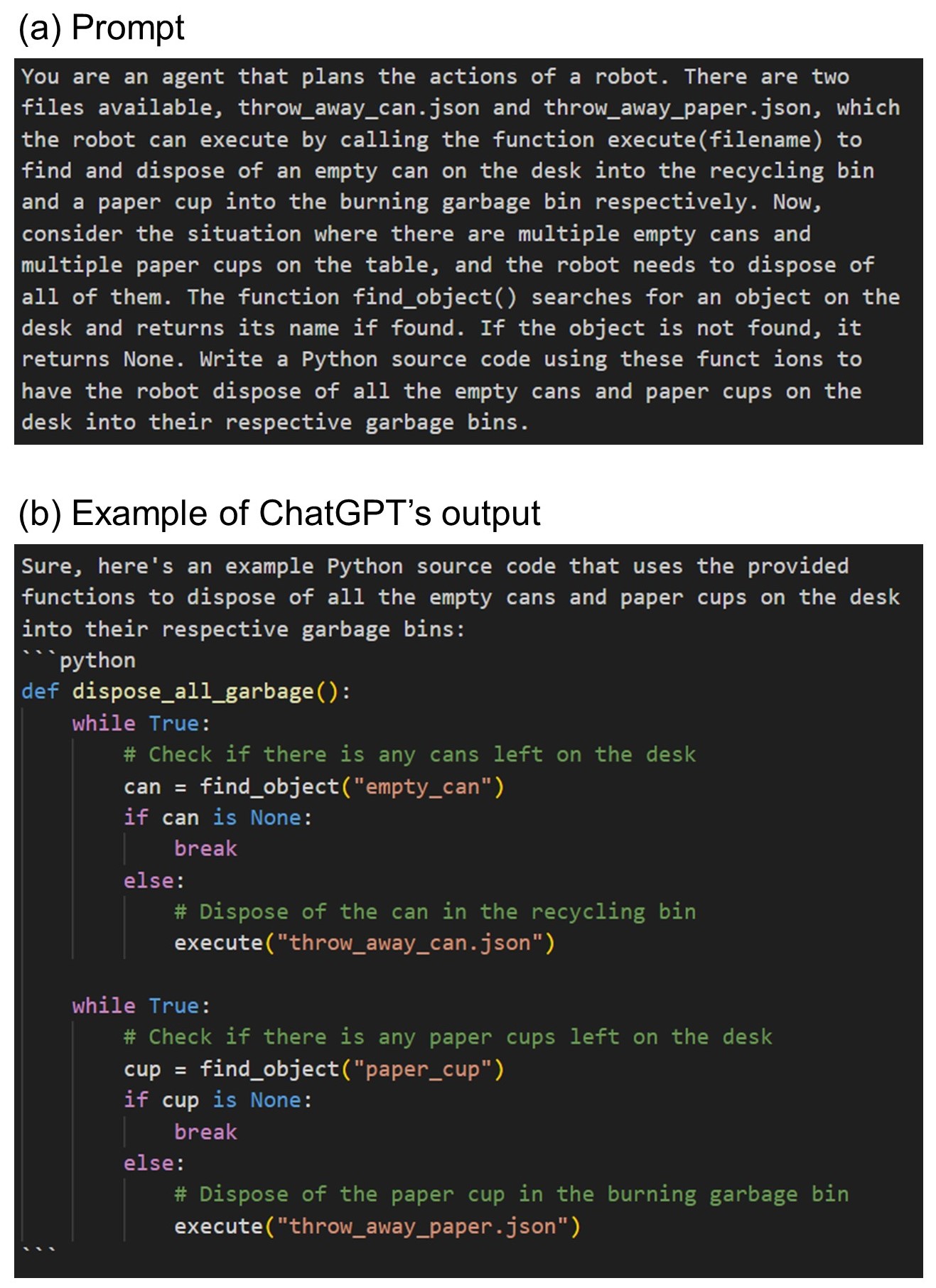}
  \caption{
  An example of using separate ChatGPT process to generate higher-level conditional branching control by reading out stored task plans.
  }
  \label{fig:logic}
\end{figure}

Some manipulations may require selecting actions based on the recognition results (e.g., disposing of a food item if it is recognized as out of date), or require repeating actions until certain conditions are met (e.g., wiping a table until it is spotless). It is known that LLMs can generate programs that include conditional branching~\cite{chen2021evaluating}. It has also been suggested that ChatGPT can handle conditional branching for robotic applications~\cite{vemprala2023chatgpt}. Consistent with these ideas, we confirmed that small modifications to the prompts enabled ChatGPT to generate a Python code that included conditional branching (Fig.~\ref{fig:logic_prompt}). Additionally, we verified that employing a separate ChatGPT process enables higher-level conditional branching by composing sets of task plans (Fig.~\ref{fig:logic}). These results suggest the feasibility of extending the proposed task planner to handle conditional branching.

\subsection{Collaboration of multiple arms and robots}
\begin{figure}[ht]
\centering
\includegraphics[width=0.48\textwidth]{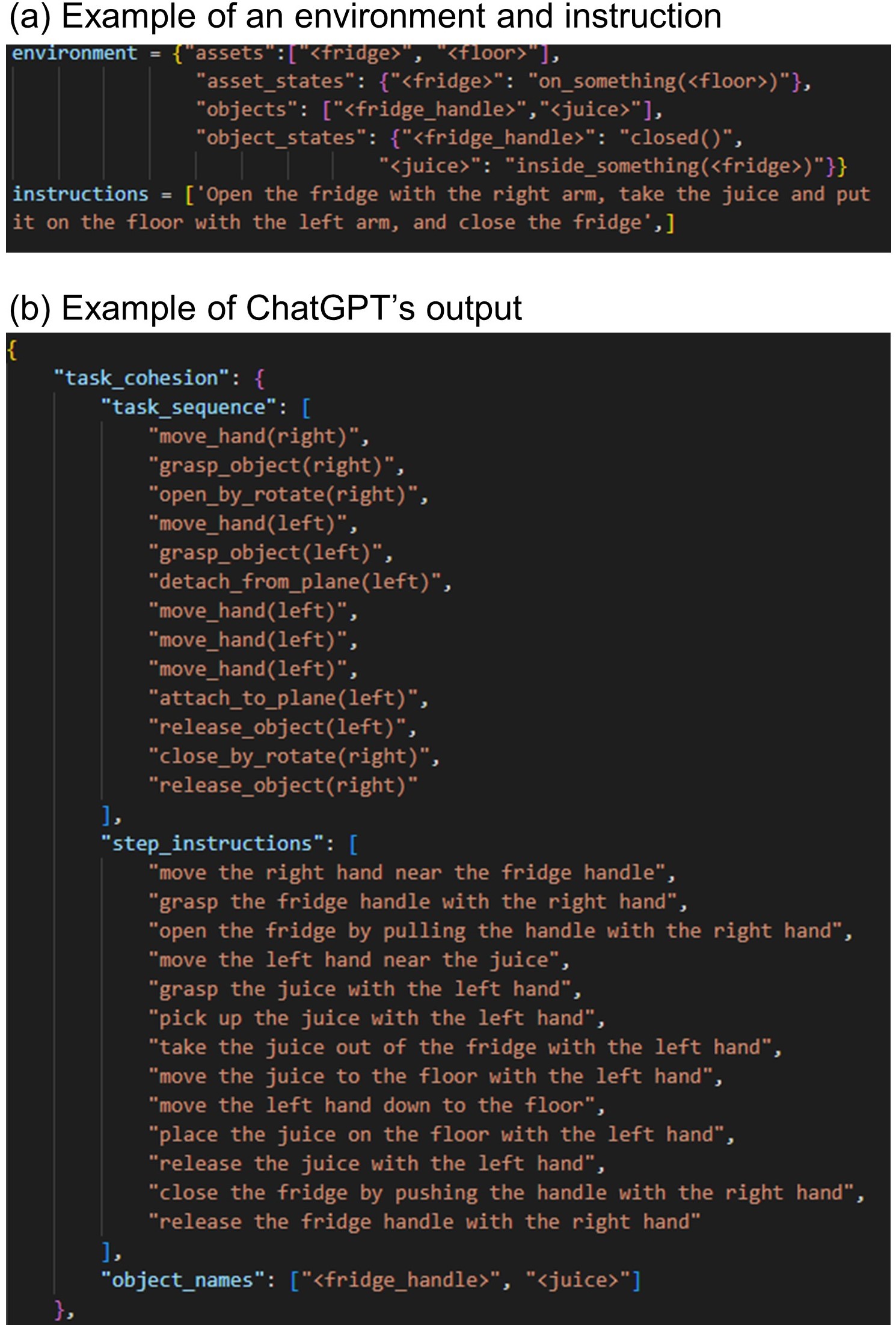}
\caption{
An example demonstrating the feasibility of ChatGPT in generating control programs that involve multiple arms or robots cooperating. Note that we included hand laterality in every function and outputted all the objects to be manipulated, as multiple objects can be handled during the grasp-manipulation-release operations of both hands. All the results, including the representation of the environment, can be found here: https://github.com/microsoft/ChatGPT-Robot-Manipulation-Prompts.
}
\label{fig:p_arms}
\end{figure}

\begin{figure}[ht]
\centering
\includegraphics[width=0.48\textwidth]{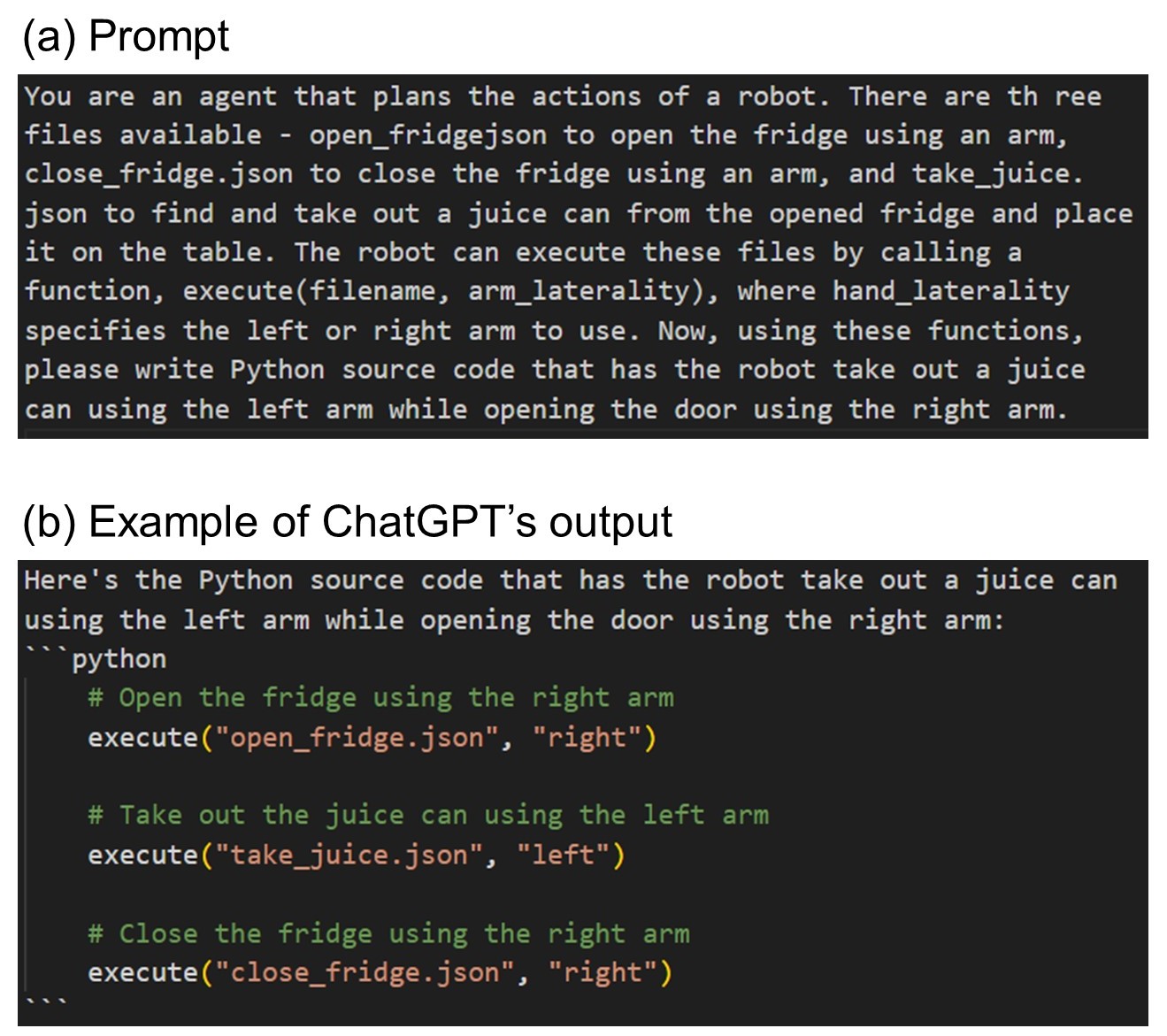}
\caption{
An example demonstrating the feasibility of ChatGPT in generating control programs that involve multiple arms or robots coordinating. This type of planning is beyond the scope of this paper.
}
\label{fig:both_arms}
\end{figure}

A robot with multiple arms may need to coordinate its arms to perform a task. We confirmed that small modifications to the prompts enabled ChatGPT to generate an action sequence involving the arms (Fig.~\ref{fig:p_arms}). 
Additionally, we verified that employing a separate ChatGPT process enables the coordination of multiple arms by composing sets of task plans (Fig.~\ref{fig:both_arms}). These results suggest the feasibility of extending the proposed task planner to handle multiple arms and robots.

\subsection{Managing environmental changes}
\begin{figure}[ht]
\centering
\includegraphics[width=0.48\textwidth]{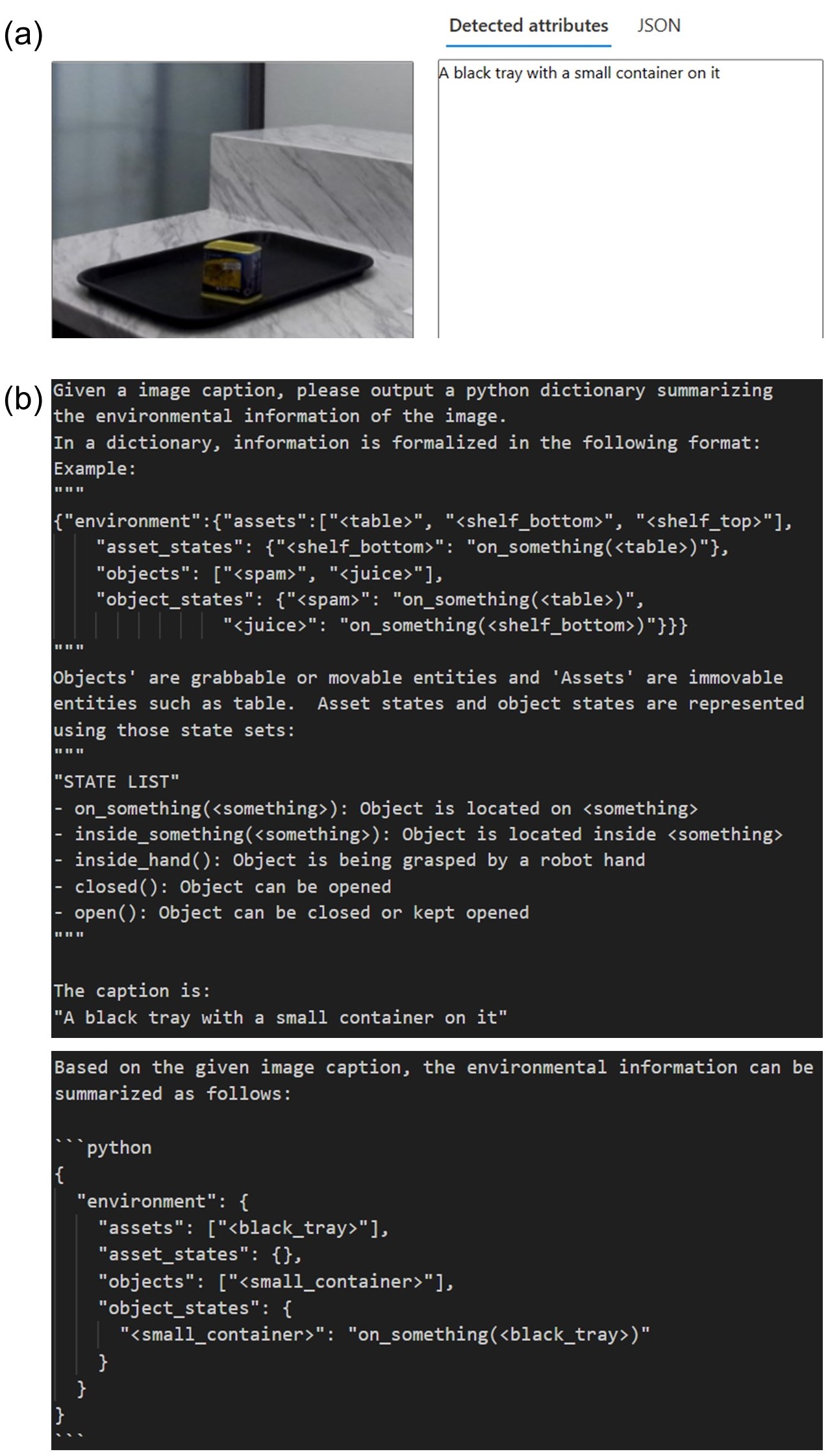}
\caption{
An example of leveraging an image description model and ChatGPT to obtain environmental information from a scene. (a) The employment of a commercially-available image description model~\cite{azure-services} to gain symbolic comprehension of the environment. (b) Utilizing ChatGPT to format the information derived from the image description.
}
\label{fig:environment}
\end{figure}

One unique aspect of our approach is that we explicitly handle changes in environmental information by incorporating it as part of the input to and output of ChatGPT, respectively. In the context of Minsky's frame theory~\cite{minsky1974framework}, environmental information can serve as ``frames'' that guide ChatGPT in selecting the most appropriate plan among a multitude of task planning options. Moreover, enabling ChatGPT to be aware of environmental information may enhance its ability to output consistent task plans~\cite{gramopadhye2022generating}. However, a limitation of this approach is the necessity to prepare environmental information, specifically for the initial instance of task planning (Fig.~\ref{fig:p_structure}). In future studies, we aim to explore a separate ChatGPT process to prepare this information based on a symbolic scene understanding given either by a vision encoder or through human explanation (Fig.~\ref{fig:environment}).

Additionally, our current approach assumes static environments, where changes are attributed solely to the robot's actions, and the environment remains consistent from task planning to execution. However, real-world scenarios frequently involve dynamic changes, such as the movement, introduction, or disappearance of objects, including people. Addressing such dynamic environments in task planning is an important direction for future research.

\subsection{Connection with vision systems and robot controllers}
Among recent experimental attempts that used ChatGPT for task planning, our work is unique in its focus on the generation of robot action sequences, addressing the ``what-to-do'' aspect, and we consciously avoid redundant language instructions related to visual and physical parameters, such as how to grab~\cite{wake2020grasp, wake2023text, saito2021contact}, where to focus~\cite{wake2020verbal}, and what posture to adopt~\cite{wake2020learning, sasabuchi2020task}, which are part of the ``how-to-do'' aspect.
Both types of information are vital for robot operation~\cite{ikeuch2023applying}, yet the ``how-to-do'' aspect is often more effectively demonstrated visually. Therefore, our design approach is such that the ``what-to-do'' is obtained by a vision system or a robot system following task planning, which is outside the scope of this paper.

\begin{figure}[ht]
\centering
\includegraphics[width=0.48\textwidth]{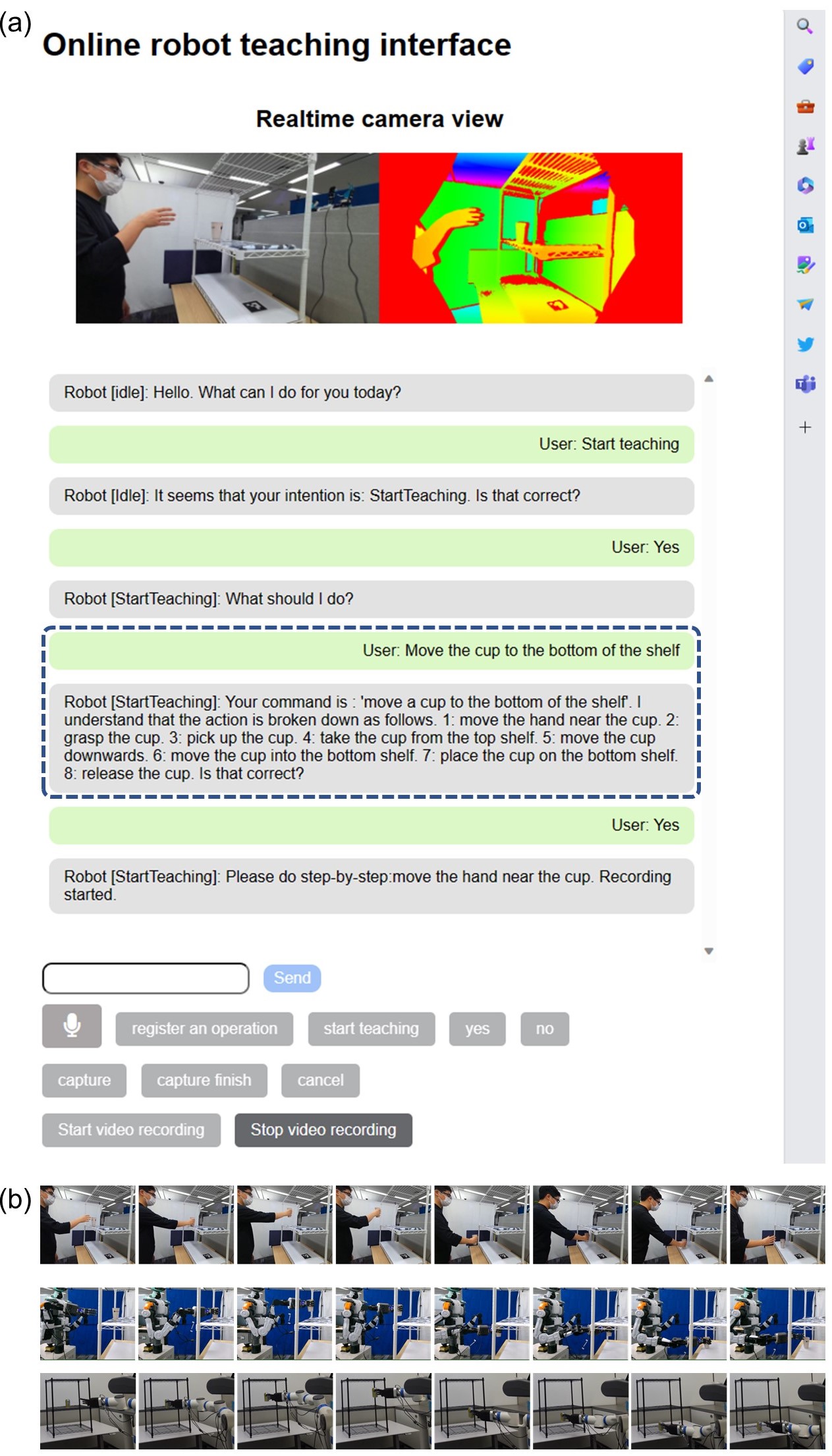}
\caption{
An example of integrating the proposed ChatGPT-empowered task planner into a robot teaching system. (a) A teaching system that incorporates the task planner (indicated by the dashed box). Following task planning, the system asks the user to visually demonstrate the tasks in a step-by-step manner. ``How-to-do'' parameters are then extracted from this visual demonstration. (b) (Top) The step-by-step demonstration corresponding to the planned tasks. (Middle and Bottom) Execution of the tasks by two different types of robot hardware.
}
\label{fig:lfo}
\end{figure}

As part of our efforts to develop a realistic robotic operation system, we have integrated our proposed task planner with a learning-from-observation system (Fig.~\ref{fig:lfo}) incorporating a speech interface\cite{wake2019enhancing, jaroslavceva2022robot}, a visual teaching interface~\cite{wake2022interactive}, a reusable robot skill library~\cite{takamatsu2022learning, saito2022task}, and a simulator~\cite{sasabuchi2023task}. 
The code for the teaching system is available at: \url{https://github.com/microsoft/cohesion-based-robot-teaching-interface}.
For reference, details of the robotic system---including how the output of ChatGPT are specifically translated into robot actions that are quantitatively controlled, how the system handles errors or unanticipated situations, and the timing for user feedback within the overall system---are provided in Appendix~\ref{appendix_robot}.

\section{Methodological considerations}
\subsection{Token limit}
Our proposed prompts aim to estimate the post-operation environment as a hint for subsequent task planning. This approach can alleviate the impact of the token limit imposed on ChatGPT, as it reduces the burden of maintaining lengthy conversation histories for multi-step instructions. However, the issue of the token limit is not completely eliminated, and it might affect the scalability of the system.  

For example, if many actions need to be explained in detail, or if an environmental description becomes lengthy as the result of including information about numerous objects, the prompts may use a significant portion of the total available tokens for the entire conversation. In such cases, one might need to accept the system limitations imposed by the token limit and adapt a strategy accordingly. This could involve simplifying the environmental descriptions or reducing the number of defined actions, in accordance with the specific scenarios being addressed (e.g., kitchen tasks, bedroom tasks, or factory assembly tasks). 

Furthermore, if a long action step is anticipated from an instruction, the need for user feedback in adjusting the output of ChatGPT may lead to increased consumption of available tokens in the conversation. In such situations, truncation of the conversation to accommodate the token limit could result in the loss of human intent included in earlier feedback (see Fig.~\ref{fig:p_structure}). This means that the token limit imposes certain restrictions on the length of actions that can be taught in a single instruction. If an instruction is likely to result in a long action sequence, the instruction may need to be broken down into smaller segments. This could help the task planning of ChatGPT, and thus reduce the amount of required feedback.

\subsection{Optimal prompting strategies}
Through our experiment with VirtualHome, we identified two failure patterns in ChatGPT: incorrect verb selection and omission of necessary steps.

Incorrect verb selection may be partially attributable to the naming conventions used for actions. In the experiment, we adhered to the original action names in VirtualHome, such as ``Put'' (an action of placing an object on another object) and ``PutIn'' (an action of placing an object inside a container with a door, such as a microwave). While these terms denote distinct actions, their similar names could potentially lead to confusion. To verify this hypothesis, we conducted a follow-up experiment where we renamed the actions to ``PutSurface'' and ``PutContainerWithDoor,'' respectively, to reflect their definitions more accurately. This modification led to a reduction in this type of error (data not shown\footnote{The results of the follow-up experiments can be found here: \url{https://github.com/microsoft/ChatGPT-Robot-Manipulation-Prompts}}), underscoring the importance of precise action naming when instructing ChatGPT—--a finding that aligns with prior research~\cite{vemprala2023chatgpt}.

Omission of necessary steps may partially stem from ChatGPT's difficulty in interpreting the granularity of defined actions. The granularity of actions implied by language is often ambiguous. For example, an instruction such as ``Put food in the microwave'' could be perceived either as a single action or a sequence of more detailed actions, such as ``open the microwave, put food in it, close it.'' Despite having provided action definitions in the prompt, the inherent ambiguity in language may lead to the omission of necessary actions in task planning. To address this failure pattern, providing more examples could effectively guide ChatGPT to decompose intended actions at the desired level of granularity. Although our initial experiments with VirtualHome included one pick-and-place example in the prompt, a follow-up experiment confirmed that including an example of placing food in a microwave reduced the occurrence of step omission---specifically, the omission of opening and closing actions (data not shown\footnotemark[2]).

\subsection{Variations in the expression of instructions}\label{instructions_variation}
In our experiments, the instructions used were relatively explicit expressions that directly specified the actions to be performed. While the use of LLMs in processing such expressions might not yield significantly greater benefits compared to conventional machine learning techniques (e.g., \cite{tellex2020robots}) for text processing, one advantage of LLMs is suggested to lie in their ability to handle high-level texts~\cite{ye2023improved, cai2023does}. Therefore, as a follow-up experiment, we adjusted our instructions to focus on the desired outcomes and objectives of the operation (e.g., ``Let's watch TV.'' instead of ``Turn on the TV.,'' see Table~\ref{table:scenarioswithactions} in Appendix~\ref{appendix} for details). As a result, we observed a performance level consistent with that of the original instructions (Table~\ref{table:result_vh_highlevel} in Appendix~\ref{appendix}). Additionally, we tested the task planner with various instructions that contained similar intent but were worded differently for a given scenario (e.g., ``Take the book from the table and put it on the bookshelf.'' and ``Grab the book from the table and place it on the bookshelf.''), and confirmed that the performance level remained consistent across variations in instruction (data not shown\footnotemark[2]).

These results do not imply that our task planner can accommodate any forms of variation in expression, yet suggest its effectiveness to a reasonable extent. Nevertheless, we reiterate that the robustness and soundness of the proposed task planning can be supported more by the functionality allowing for necessary adjustments through user feedback, rather than the performance of single-shot task planning. 
\subsection{Data privacy and security}
In response to emerging concerns regarding data privacy and security, careful data handling is crucial for systems that rely on LLMs. To address this issue, we operate ChatGPT via Azure OpenAI, a service provided by Microsoft. This service enables us to handle data in compliance with various legal regulations and standards related to data security and privacy, ensuring the proper protection of user instructions and information~\cite{AzureOpenAISecurity}. Furthermore, Azure OpenAI includes features for content filtering and abuse monitoring, which aid in mitigating risks associated with misuse. Thus, we believe that our task planning system can operate while meeting industry standards for data privacy and security. However, it is essential for future work to continually assess and improve these protective measures as our understanding of potential risks evolves.

\subsection{Future directions}
Among the pioneering studies for task planning from natural language, a significant advantage of utilizing the most recent LLMs is their adaptability to various operational settings through few-shot learning and user feedback. These functionalities not only remove the need for extensive data collection or model retraining but also enable user adjustments, thereby facilitating safe and robust task planning.

While we use ChatGPT as an example of such an LLM, these capabilities are not confined to any specific model. The ability to perform few-shot learning is considered a result of increased model sizes~\cite{brown2020language} and extended training on large datasets~\cite{hoffmann2022training}. Furthermore, the capacity to effectively accommodate user feedback could be partially attributed to learning methods that align model behavior with human intent, known as reinforcement learning from human feedback~\cite{ouyang2022training}. In fact, other models that utilize similar training techniques, such as GPT-4~\cite{OpenAIgpt4} and Llama2-chat~\cite{touvron2023llama}, have been reported to possess these features. Future research will explore whether other models can yield results comparable to those found in this study when applied to task planning.

Regarding the adjustment capability, our experiments suggested that the output of ChatGPT can be adjusted through a reasonable amount of feedback. ChatGPT's ability to reflect the semantic content of user feedback provides a means for users to convey their intentions to the system. Thus, we consider that this aspect contributes to the foundation of a user-friendly system. However, this study did not delve into how this adjustment capability directly contributes to user-friendliness of the system. Future research areas include user studies focusing on usability and comparisons with other adjustment methods, such as directly editing the output of ChatGPT.

\section{Conclusion}
This paper presented a practical application of OpenAI's ChatGPT for translating multi-step instructions into executable robot actions. We designed input prompts to meet the common requirements in practical applications, specifically encouraging ChatGPT to output a sequence of robot actions in a readable format and explicitly handle the environmental information before and after executing the actions. Through experiments, we tested the effectiveness of our proposed prompts in various environments. Additionally, we observed that the conversational capability of ChatGPT allows users to adjust the output through natural-language feedback, which is crucial for safe and robust task planning. Our prompts and source code are open-source and publicly available. We hope that this study will provide practical resources to the robotics research community and inspire further developments in this research area.

\section*{Acknowledgment}
This study was conceptualized, conducted, and written by the authors, and an AI (OpenAI's GPT-4 model) was used for proofreading.

\bibliographystyle{ieeetr}
\bibliography{bib}

\appendices
\section{Detailed prompt for defining robot actions}\label{prompt}
Fig.~\ref{fig:action_all} provides the unabridged prompt that is exemplified in Section~\ref{action_definition}. It includes the list of robot actions and their definitions.
\begin{figure}[ht]
\begin{mdframed}[backgroundcolor=black]
\begin{flushleft}
\color[rgb]{0.7,0.7,0.7}\scriptsize
Necessary and sufficient robot actions are defined as follows:\\
\textquotedbl\textquotedbl\textquotedbl\\
"ROBOT ACTION LIST"\\
- move\_hand(): Move the robot hand from one position to another with/without grasping an object.\\
- grasp\_object(): Grab an object.\\
- release\_object(): Release an object in the robot hand.\\
- move\_object(): Move the object grabbed by the robot hand from one position to another. move\_object() is allowed only when the object is not physically constrained by the environment. For example, if the robot hand moves an object on the floor to another location, move\_object() is not allowed because the object is constrained by the floor.\\
- detach\_from\_plane(): This action can only be performed if an object is grabbed. Move the grabbed object from a state in which it is constrained by a plane to a state in which it is not constrained by any plane. For example, detach\_from\_plane() is used when a robot hand picks up an object on a table.\\
- attach\_to\_plane(): This action can only be performed if an object is grabbed. The opposite operation of detach\_from\_plane().\\
- open\_by\_rotate(): This action can only be performed if an object is grabbed. Open something by rotating an object that is rotationally constrained by its environment along its rotation. For example, when opening a refrigerator, the refrigerator handle makes this motion. Also, when opening the lid of a plastic bottle, the lid makes this motion.\\
- adjust\_by\_rotate(): This action can only be performed if an object is grabbed. Rotate an object that is rotationally constrained by its environment along its rotation. For example, when adjusting the temperature of a refrigerator, the temperature knob makes this motion.\\
- close\_by\_rotate(): This action can only be performed if an object is grabbed. The opposite operation of open\_by\_rotate().\\
- open\_by\_slide(): This action can only be performed if an object is grabbed. Moves an object that is translationally constrained in two axes from its environment along one unbounded axis. For example, when opening a sliding door or drawer, the handle makes this movement.\\
- adjust\_by\_slide(): This action can only be performed if an object is grabbed. Slide an object that is translationally constrained in two axes from its environment along one unbounded axis. For example, when widen the gap between a sliding door and the wall, the handle makes this movement.\\
- close\_by\_slide(): This action can only be performed if an object is grabbed. The opposite operation of open\_by\_slide().\\
- wipe\_on\_plane(): This action can only be performed if an object is grabbed. Move an object landing on a plane along two axes along that plane. For example, when wiping a window with a sponge, the sponge makes this motion.\\
\textquotedbl\textquotedbl\textquotedbl\\
\end{flushleft}
\end{mdframed}
\caption{
The prompt explaining the robotic functions.
}
\label{fig:action_all}
\end{figure}

\section{Supplementary information for VirtualHome Experiment}\label{appendix}
This section provides supplementary information for the VirtualHome experiment discussed in Section~\ref{virtualhome}. Table~\ref{tab:human_action_list} displays a list of pre-defined atomic actions in VirtualHome, which represent the smallest units of action. Table \ref{table:scenarioswithactions} illustrates fourteen scenarios used for the experiment. The ``Textual instruction'' column indicates instructions that were fed into the task planner. The ``Action sequence'' column shows the manually identified action sequences to achieve the scenarios. The ``Higher-level textual instruction'' column displays instructions that have been adjusted to emphasize desired outcomes and objectives of the operation. These instructions were used in a follow-up experiment that tested ChatGPT's ability to understand and respond to higher-level texts (See Section~\ref{instructions_variation}). Table~\ref{table:result_vh_highlevel} shows the results. 

\begin{table}[ht]
\caption{The action list defined for the experiment in Section~\ref{virtualhome}}
\centering
\begin{tabularx}{0.49\textwidth}{|r|X|}
\hline
\textbf{Action} & \textbf{Description} \\
\hline
Walktowards(arg1) & Walks some distance towards a room or object. \\
\hline
Grab(arg1) & Grabs an object. \\
\hline
Open(arg1) & Opens an object. \\
\hline
Close(arg1) & Closes an object. \\
\hline
Put(arg1, arg2) & Puts an object on another object. \\
\hline
PutIn(arg1, arg2) & Puts an object inside another container. \\
\hline
SwitchOn(arg1) & Turns an object on. \\
\hline
SwitchOff(arg1) & Turns an object off. \\
\hline
Drink(arg1) & Drinks from an object. \\
\hline
\end{tabularx}
\label{tab:human_action_list}
\end{table}

\begin{table*}[ht]
\caption{The list of scenarios and their action sequences used in the experiment}
\centering
\begin{tabularx}{\textwidth}{|r|X|l|X|}
\hline
\textbf{Scenario} & \textbf{Textual instruction} & \textbf{Action sequence} & \textbf{Higher-level textual instruction} \\ 
\hline
Scenario1 & Take the bread from the toaster on the kitchen counter and place it on the plate on the table. & \makecell{WalkTowards(toaster), \\ Grab(breadslice), \\ WalkTowards(kitchentable), \\ Put(breadslice, plate)} & Serve the toast on the table. \\
\hline
Scenario2 & Take the frying pan from the counter and place it in the sink. & \makecell{WalkTowards(stove), \\ Grab(fryingpan), \\ WalkTowards(sink), \\ Put(fryingpan, sink)} & Put away the frying pan into the sink. \\
\hline
Scenario3 & Take the pie from the table and put it in the microwave. & \makecell{WalkTowards(kitchentable), \\ Grab(pie), \\ WalkTowards(microwave), \\ Open(microwave), \\ Putin(pie, microwave), \\ Close(microwave), \\ SwitchOn(microwave)} & Heat up the pie using the microwave. \\
\hline
Scenario4 & Take the condiment shaker from the bookshelf and place it on the table. & \makecell{WalkTowards(bookshelf), \\ Grab(condimentshaker), \\ WalkTowards(kitchentable), \\ Put(condimentshaker, kitchentable)} & Set out the condiment shaker on the table. \\
\hline
Scenario5 & Take the book from the table and put it on the bookshelf. & \makecell{WalkTowards(kitchentable), \\ Grab(book), \\ WalkTowards(bookshelf), \\ Put(book, bookshelf)} & Store the book on the shelf. \\
\hline
Scenario6 & Take the water glass from the table and drink from it. & \makecell{WalkTowards(kitchentable), \\ Grab(waterglass), \\ Drink(waterglass), \\ Put(waterglass, kitchentable)} & Drink the water from the grass. \\
\hline
Scenario7 & Take the salmon on top of the microwave and put it in the fridge. & \makecell{WalkTowards(microwave), \\ Grab(salmon), \\ WalkTowards(fridge), \\ Open(fridge), \\ Putin(salmon, fridge), \\ Close(fridge)} & Chill the salmon in the fridge. \\
\hline
Scenario8 & Turn on the TV. & \makecell{WalkTowards(tvstand), \\ SwitchOn(tv)} & Let's watch TV. \\
\hline
Scenario9 & Put a plate that is on the table into the sink. & \makecell{WalkTowards(kitchentable), \\ Grab(plate), \\ WalkTowards(sink), \\ Put(plate, sink)} & Clear away that plate into the sink. \\
\hline
Scenario10 & Take the pie on the table and warm it using the stove. & \makecell{WalkTowards(kitchentable), \\ Grab(pie), \\ WalkTowards(stove), \\ Open(stove), \\ Putin(pie, stove), \\ Close(stove) \\ SwitchOn(stove)} & Warm the pie using the stove. \\
\hline
Scenario11 & Put the sponge in the sink and wet it with water. & \makecell{WalkTowards(kitchencounter), \\ Grab(washingsponge), \\ WalkTowards(sink), \\ Put(washingsponge, sink), \\ SwitchOn(faucet)} & Wet the sponge. \\
\hline
Scenario12 & Take the knife from the table and move it to another place on the table. & \makecell{WalkTowards(kitchentable), \\ Grab(cutleryknife), \\ WalkTowards(kitchentable), \\ Put(cutleryknife, kitchentable)} & Find another place for the knife. \\
\hline
Scenario13 & Take the plate from the table and move it to another place on the table. & \makecell{WalkTowards(kitchentable), \\ Grab(plate), \\ WalkTowards(kitchentable), \\ Put(plate, kitchentable)} & Reposition that plate. \\
\hline
Scenario14 & Take the condiment bottle from the bookshelf and put it on the table. & \makecell{WalkTowards(bookshelf), \\ Grab(condimentbottle), \\ WalkTowards(kitchentable), \\ Put(condimentbottle, kitchentable)} & Place the condiment bottle on the table. \\
\hline
\end{tabularx}
\label{table:scenarioswithactions}
\end{table*}

\begin{table*}[ht]
\caption{Executability of the output action sequence across trials (Higher-level textual instruction). ``1'' indicates success, and ``0'' indicates failure.}
\centering
\begin{tabular*}{0.7\textwidth}{@{\extracolsep{\fill}}|c|c|c|c|c|c|c|c|c|c|c|c|c|c|c|}
\hline
Scenario & 1 & 2 & 3 & 4 & 5 & 6 & 7 & 8 & 9 & 10 & 11 & 12 & 13 & 14 \\
\hline
Trial 1 & 0 & 0 & 0 & 0 & 1 & 1 & 0 & 1 & 0 & 0 & 0 & 0 & 0 & 1 \\
Trial 2 & 1 & 0 & 0 & 0 & 1 & 1 & 0 & 1 & 0 & 0 & 0 & 1 & 1 & 1 \\
Trial 3 & 1 & 0 & 0 & 0 & 1 & 1 & 0 & 1 & 0 & 0 & 0 & 1 & 1 & 1 \\
Trial 4 & 0 & 0 & 0 & 0 & 1 & 1 & 0 & 1 & 0 & 0 & 0 & 0 & 0 & 1 \\
Trial 5 & 0 & 0 & 0 & 0 & 1 & 1 & 0 & 1 & 0 & 0 & 0 & 1 & 1 & 1 \\
\hline
\end{tabular*}
\label{table:result_vh_highlevel}
\end{table*}

\section{An example of a robot system expanding the proposed task planner}\label{appendix_robot}

In this supplementary section, we present an overview of our in-house robot teaching system, which serves as an illustrative example of the proposed task planner. Importantly, our system is designed for use under the guidance of experts familiar with robot operations and action definitions, rather than being an automatic solution for non-experts. The system's main objective is to simplify the robot teaching process, eliminating the need for complex coding by incorporating a method of robot programming that uses multimodal demonstrations and language feedback.

\textbf{Prerequisite}: The representation of the initial environment and the sequence of instructions are assumed to have been prepared manually. 

\begin{figure}[ht]
\centering
\includegraphics[width=0.48\textwidth]{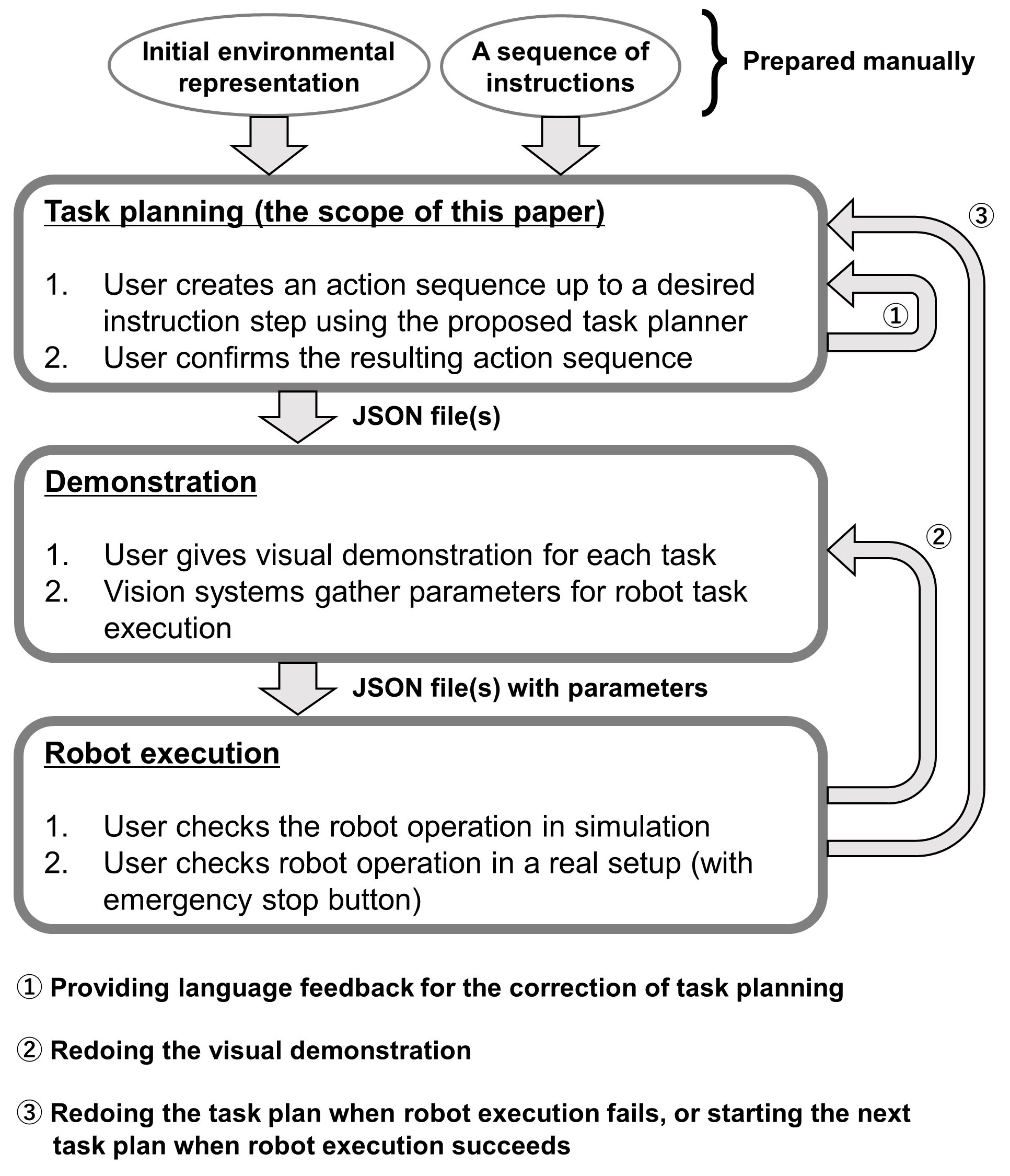}
\caption{Overview of the robot teaching system that integrates the proposed task planner. The process involves three main steps: Task planning, where the user employs the task planner to create an action sequence and adjusts the result through feedback as necessary; Demonstration, where the user visually demonstrates the action sequence to provide information needed for robot operation; and Robot Execution, where the action sequence is first simulated and then tested. If any step fails or shows deficiencies, the previous steps can be revisited as necessary.}
\label{fig:overview_robotsystem}
\end{figure}

The robot teaching system operates through the following three steps (Fig.~\ref{fig:overview_robotsystem}):
\begin{enumerate}
\item \textbf{Task planning (the scope of this paper)}: The user creates a task plan up to a desired instruction step using the proposed task planner. If any deficiencies are found in the output sequence, the user can provide feedback to the task planner as necessary.
\item \textbf{Demonstration}: The user visually demonstrates the action sequence to provide information needed for robot operation. Specifically, the system asks the user to demonstrate each task step-by-step in front of an RGB-D camera. The vision system then analyzes the visual demonstration and extracts the parameters needed for the robot to perform each task.
\item \textbf{Robot execution}: The user first simulates the action sequence and checks the results. The simulation environment is designed to replicate the actual one~\cite{sasabuchi2023task}. If execution fails or leads to an unexpected result, the task planning and demonstration steps are revisited as necessary. Only when safe operation is confirmed in the simulation does the user test the action sequence in the real setup. For safety, the robot operation is tested under a condition where the user can press the robot's emergency stop switch at any time.
\end{enumerate}

Examples of the parameters required for the robot to execute each task are provided in Table~\ref{tab:robotactionsandparameters}. In step 2, the vision system identifies the parameters listed in the table's second column by utilizing third-party pose recognizers and object recognizers. For example, parameters such as the center of rotation, rotation axis, and rotation radius are estimated from the hand's trajectory. To represent arm postures, we have prepared 26 unit vectors that indicate 3D directions. These vectors are used to represent the discrete orientation of upper and lower arms. Specifically, we choose the vector closest to the direction in which each part of the arm is pointing, using it to represent that part's direction~\cite{wake2020learning}. More detailed methods for acquiring parameters are explained in other papers~\cite{wake2020learning, wake2020verbal}.

In Step 3, besides the aforementioned parameters, the robot system controls the robot using data from an RGB-D camera and force sensors mounted on it (see the third column of Table~\ref{tab:robotactionsandparameters}). It is assumed that the environment at the start of the demonstration and execution will be identical to the extent that neither the action sequence nor the discrete representation of the posture would be affected. The robot's vision system looks for objects again during execution and corrects slight misalignments. For some tasks, the value of the force sensors attached to and near the end effector is used as force feedback. 

The robot system computes the robot's physical movements for tasks such as move\_hand(), move\_object(), and release\_object() by using inverse kinematics with postural constraint~\cite{sasabuchi2020task}, following the parameters recognized by the vision system. For other tasks, the movements are computed by pre-trained reinforcement learning policies~\cite{takamatsu2022learning, saito2022task}. Notably, the post-operation environment output by ChatGPT is used only as a hint for subsequent task planning but is not used during robot execution in our robot system. It is worth reiterating that task definitions and execution methods differ depending on the design philosophy, and this table merely illustrates one example of the implementation.

\begin{table*}[ht]
\caption{Robot actions and the parameters obtained from visual demonstrations and during on-site robot execution}
\centering
\begin{tabularx}{\textwidth}{|l|X|X|}
\hline
\textbf{Robot action} & \textbf{Parameters obtained from corresponding visual demonstrations (RGB-D images)} & \textbf{Parameters obtained during on-site robot execution (RGB-D images and force sensors)} \\
\hline
move\_hand() & 
\begin{itemize}
  \item The 3D positions of the manipulating hand at the first and the last video frames.
  \item Whether the manipulating hand is left or right.
  \item The arm postures at the first and the last video frames.
\end{itemize} & NA \\
\hline
grasp\_object() &
\begin{itemize}
  \item The 3D position of the object.
  \item Whether the manipulating hand is left or right.
  \item The arm postures at the first and the last video frames.
  \item The approach direction of the hand to the object.
  \item The grasp type according to a grasping taxonomy.
\end{itemize} &
\begin{itemize}
  \item The 3D position of the contact points where the robot's end effectors should engage, as determined by the grasp type.
  \item The value of the force sensor attached to the robot's end effectors.
\end{itemize} \\
\hline
release\_object() &
\begin{itemize}
  \item The retreat direction of the hand from the object.
  \item Whether the manipulating hand is left or right.
  \item The arm postures at the first and the last video frames.
\end{itemize} & NA \\
\hline
move\_object() &
\begin{itemize}
  \item The 3D positions of the manipulating hand at the first and the last video frames.
  \item Whether the manipulating hand is left or right.
  \item The arm postures at the first and the last video frames.
\end{itemize} & NA \\
\hline
detach\_from\_plane() &
\begin{itemize}
  \item The displacement when the object in hand detaches from the plane.
  \item The 3D position of the manipulating hand at the last video frame.
  \item Whether the manipulating hand is left or right.
  \item The arm postures at the first and the last video frames.
\end{itemize} &
\begin{itemize}
  \item The value of the force sensor attached to the wrist area near the robot's end effector.
\end{itemize} \\
\hline
attach\_to\_plane() &
\begin{itemize}
  \item The displacement when the object in hand attaches to the plane.
  \item The 3D position of the manipulating hand at the first video frame.
  \item Whether the manipulating hand is left or right.
  \item The arm postures at the first and the last video frames.
\end{itemize} &
\begin{itemize}
  \item The value of the force sensor attached to the wrist area near the robot's end effector.
\end{itemize} \\
\hline
\makecell{open\_by\_rotate(), \\ adjust\_by\_rotate(), \\ close\_by\_rotate()} &
\begin{itemize}
  \item The 3D positions of the manipulating hand across all the video frames.
  \item The direction of the rotation axis.
  \item The position of the rotation center.
  \item The angle rotation.
  \item Whether the manipulating hand is left or right.
  \item The arm postures at the first and the last video frames.
\end{itemize} & 
\begin{itemize}
  \item The value of the force sensor attached to the wrist area near the robot's end effector.
\end{itemize} \\
\hline
\makecell{open\_by\_slide(), \\ adjust\_by\_slide(), \\ close\_by\_slide()} &
\begin{itemize}
  \item The 3D positions of the manipulating hand across all the video frames.
  \item The displacement of the sliding motion.
  \item Whether the manipulating hand is left or right.
  \item The arm postures at the first and the last video frames.
\end{itemize} &
\begin{itemize}
  \item The value of the force sensor attached to the wrist area near the robot's end effector.
\end{itemize} \\
\hline
wipe\_on\_plane() &
\begin{itemize}
  \item The 3D positions of the manipulating hand across all the video frames.
  \item The axis that is vertical to the wiping plane.
  \item Whether the manipulating hand is left or right.
  \item The arm postures at the first and the last video frames.
\end{itemize} &
\begin{itemize}
  \item The value of the force sensor attached to the wrist area near the robot's end effector.
\end{itemize} \\
\hline
\end{tabularx}
\label{tab:robotactionsandparameters}
\end{table*}

\end{document}